\documentclass[preprint,3p,times,twocolumn]{elsarticle} 
\usepackage[pagebackref=true,breaklinks=true,letterpaper=true,colorlinks,bookmarks=false]{hyperref}
\usepackage{nth}
\usepackage{nicefrac}
\usepackage{relsize}
\usepackage{graphicx}
\usepackage{booktabs}
\usepackage{multirow}
\usepackage{pbox}
\usepackage{amsmath} 
\usepackage{amssymb}
\usepackage{xspace} 
\usepackage{bm}
\usepackage{enumerate}
\usepackage[table]{xcolor}
\usepackage{enumitem}
\usepackage{array}
\usepackage[font=normal,skip=5pt]{caption}
\usepackage{subcaption}
\usepackage{dblfloatfix}
\usepackage[linesnumbered,ruled]{algorithm2e}
\usepackage{algpseudocode}
\usepackage{bm}
\usepackage{dsfont}

\newcommand{\etal}{\textit{et al. }}
\newcommand{\eg}{e.g.}
\newcommand{\ie}{i.e.}
\usepackage{longtable}
\usepackage{epstopdf}
\usepackage[export]{adjustbox}
\usepackage{color}
\newcolumntype{P}[1]{>{\centering\arraybackslash}p{#1}}
\journal{Pattern Recognition}
\begin{document}
	
	\begin{frontmatter}
		
		\title{Curriculum Learning of Visual Attribute Clusters \\for Multi-Task Classification }
		
		\author[uh]{Nikolaos Sarafianos}
		\ead{nsarafia@central.uh.edu}
		\author[ncsr]{Theodore Giannakopoulos}
		\author[uoi]{Christophoros Nikou}
		\author[uh]{Ioannis A. Kakadiaris}
		\address[uh]{Computational Biomedicine Lab, Department of Computer Science\\
			University of Houston, 4800 Calhoun Rd. Houston, TX 77004}
		\address[ncsr]{National Center of Scientific Research Demokritos, Athens GR 15310, Greece}
		\address[uoi]{Department of Computer Science and Engineering, University of Ioannina, Ioannina GR 45110, Greece}
		\begin{abstract}
			Visual attributes, from simple objects (\eg, backpacks, hats) to soft-biometrics (\eg, gender, height, clothing) have proven to be a powerful representational approach for many applications such as image description and human identification. In this paper, we introduce a novel method to combine the advantages of both multi-task and curriculum learning in a visual attribute classification framework. Individual tasks are grouped after performing hierarchical clustering based on their correlation. The clusters of tasks are learned in a curriculum learning setup by transferring knowledge between clusters. The learning process within each cluster is performed in a multi-task classification setup. By leveraging the acquired knowledge, we speed-up the process and improve performance. We demonstrate the effectiveness of our method via ablation studies and a detailed analysis of the covariates, on a variety of publicly available datasets of humans standing with their full-body visible. Extensive experimentation has proven that the proposed approach boosts the performance by 4\% to 10\%.
		\end{abstract}
		
		
	\end{frontmatter}
	\section{Introduction}\label{sec:introduction} 
	\begin{flushright}
		\small\textit{Vision as reception. Vision as reflection. Vision as projection.\\
			--Bill Viola, note 1986}\\
	\end{flushright}
	When we are interested in providing a description of an object or a human, we tend to use visual attributes to accomplish this task. For example, a laptop can have a wide screen, a silver color, and a brand logo, whereas a human can be tall, female, wearing a blue t-shirt and carrying a backpack. Visual attributes in computer vision are equivalent to the adjectives in our speech. We rely on visual attributes since (i) they enhance our understanding by creating an image in our head of what this object or human looks like; (ii) they narrow down the possible related results when we want to search for a product online or when we want to provide a suspect description; (iii) they can be composed in different ways to create descriptions; (iv) they generalize well as with some fine-tuning they can be applied to recognize objects for different tasks; and (v) they are a meaningful semantic representation of objects or humans that can be understood by both computers and humans. However, effectively predicting the corresponding visual attributes of a human given an image remains a challenging task \cite{liu2017best}. In real-life scenarios, images might be of low-resolution, humans might be partially occluded in cluttered scenes, or there might be significant pose variations.  
	
	Estimating the visual attributes of humans is an important computer vision problem with applications ranging from finding missing children to virtual reality. When a child goes missing or the police is looking for a suspect, a short description is usually provided that comprises such attributes (for example, tall white male, with a black shirt wearing a hat and carrying a backpack). Thus, if we could efficiently identify which images or videos contain images of humans with such characteristics we could potentially reduce dramatically the labor and the time required to identify them \cite{fabian2014visual}. Another interesting application is the 3D reconstruction of the human body in virtual reality \cite{stanney1998human}. If we have such attribute information we can facilitate the reconstruction by providing the necessary priors. For example it is easier to reconstruct accurately the body shape of a human if we already know that it is a tall male with shorts and sunglasses than if no information is provided.
	
	\begin{figure}[t] 
		\centering
		\includegraphics[width=0.49\textwidth]{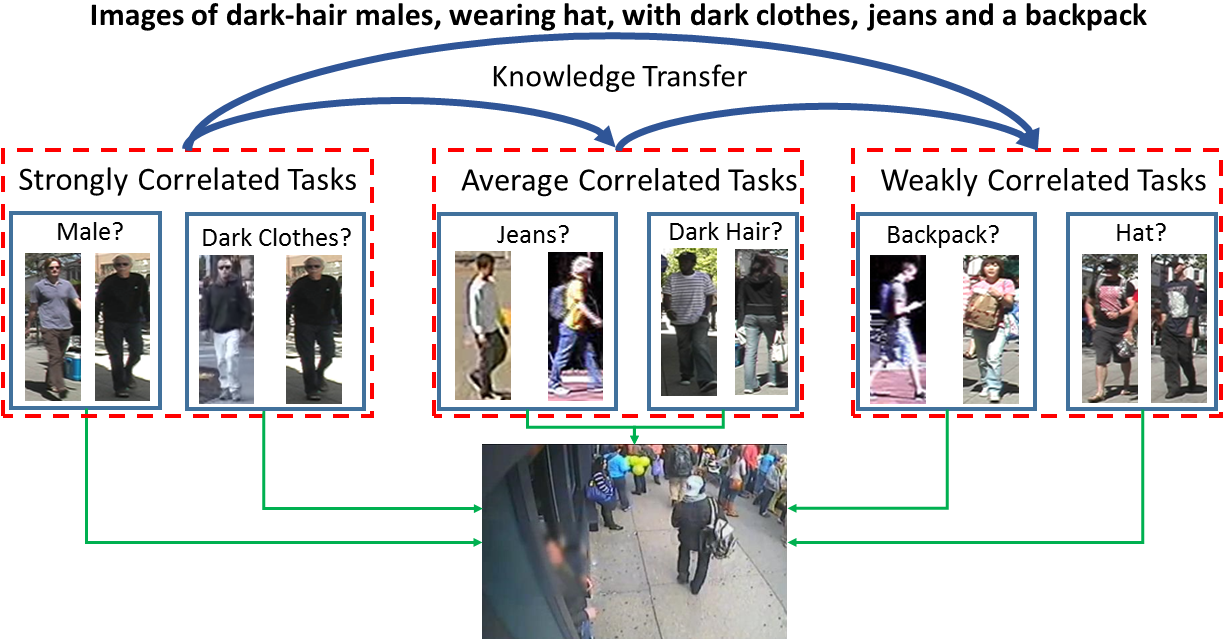}
		\caption{Curriculum learning for multi-task classification of visual attributes. Tasks are split into groups by performing hierarchical clustering which are then learned sequentially based on the cross-correlation of the attributes within each group. Flickr photo by Jeffery Scism is licensed under \href{https://creativecommons.org/licenses/by/2.0/}{CC BY}.}
		\label{fig:Boston}
	\end{figure} 
	
	In this work, we introduce CILICIA (CurrIculum Learning multItask ClassIfication Attributes) to address the problem of visual attribute classification from images of standing humans. Instead of using low-level representations, which would require extracting hand-crafted features, we propose a deep learning method to solve multiple binary classification tasks. CILICIA differentiates itself from the literature as: (i) it performs end-to-end learning by feeding a single ConvNet with the entire image of a human without making any assumptions about predefined connection between body parts and image regions; and (ii) it exploits the advantages of both multi-task and curriculum learning. Tasks are split into groups based on their labels' cross-correlation using hierarchical agglomerative clustering. The groups of tasks are learned in a curriculum learning scenario, starting with the one with the highest within-group cross-correlation and moving to the less correlated ones by transferring knowledge from the former to the latter. The tasks in each group are learned in a typical multi-task classification setup.   Parts of this publication appear in our previous work \cite{sarafianos2017curriculum}. However, in this work we have:
	\begin{itemize}
		\item Proposed an effective method to obtain the groups of tasks using hierarchical agglomerative clustering, which can be of any number and not just two groups (strongly/weakly correlated).
		\item Conducted additional experiments to analyze the covariates of the proposed approach.
		\item Benchmarked our method in an additional dataset.
		\item Demonstrated the efficacy and robustness of our method by performing ablation studies in Section~\ref{sec:ablations}.
	\end{itemize} 
	
	When Vapnik and Vashist introduced the learning using privileged information (LUPI) paradigm \cite{Vapnik_2009_15070}, they drew inspiration from human learning. They observed how significant the role of an intelligent teacher was in the learning process of a student, and proposed a machine learning framework to imitate this process. Employing privileged information from an intelligent teacher at training time has recently received significant attention from the scientific community with remarkable results in areas ranging from object recognition \cite{motiian2016information, Pechyony_2010_16768, Sharmanska_2013_16765, Wang_2015_16756} to biometrics \cite{kakadiaris2016show, vrigkas2016exploiting, sarafianos2016icpr}. 
	
	Our work also draws inspiration from the way students learn in class. First, students find it difficult to learn all tasks at once. It is usually easier for them to acquire some basic knowledge first, and then build on top of that, by learning more complicated concepts. This can be achieved by learning in a hierarchical manner, which is commonly employed in the literature \cite{yanhd, zhang2015hierarchical, zhang2017hierarchical}, or with a curriculum strategy. Curriculum learning \cite{bengio2009curriculum, jiang2014self, graves2017automated} (presenting easier examples before more complicated and learning tasks sequentially, instead of all at the same time) imitates this learning process. It has the advantage of exploiting prior knowledge to improve subsequent classification tasks but it cannot scale up to many tasks since each subsequent task has to be learned individually. However, to maximize students' understanding a curriculum might not be sufficient by itself. Students also need a teaching paradigm that can guide their learning process, especially when the task to be learned is challenging. The teaching paradigm in our method is the split of visual attribute classification tasks that need to be learned by performing hierarchical agglomerative clustering. In that way, we exploit the advantages of both multi-task and curriculum learning. First, the ConvNet learns the group of tasks with the strongest intra cross-correlation in a multi-task learning setup, and once this process is completed, the weights of the respective tasks are used as an initialization for the more diverse tasks. During the training of the more diverse tasks, the prior knowledge obtained is leveraged to improve the classification performance. An illustrative example of our method is depicted in Figure~\ref{fig:Boston}. Note that the proposed learning paradigm is not tied visual attribute classification domain and can be extended to other applications such as object recognition or \cite{hariharan2012discriminative} and domain adaptation \cite{dong2016multi}.

	In summary, this paper has the following contributions. First, we introduce CILICIA, a novel method of exploiting the advantages of both multi-task and curriculum learning by splitting tasks into groups by performing hierarchical agglomerative clustering. The tasks of each subgroup are learned in a joint manner. Thus, the proposed method learns better than learning all the tasks in a typical multi-task learning setup and converges faster than learning tasks one at a time. Second, we propose a scheme of transferring knowledge between the groups of tasks which speeds up the convergence and increases the performance. We performed extensive evaluations in three datasets of humans standing and achieved state-of-the-art results in all three of them. 
	
	The remainder of the paper is organized as follows: in Section \ref{sec:related}, a review of the related work in visual attributes, curriculum learning, and transfer learning is presented. Section \ref{sec:method} presents CILICIA, the proposed curriculum learning approach for multi-task classification of clusters of visual attributes. In Section \ref{sec:exp}, experimental results are reported, a detailed analysis of covariates is provided, and a discussion about the performance and the limitations of the proposed approach is offered. Finally, conclusions are drawn in Section \ref{sec:conc}.
	
	\section{Related Work}\label{sec:related}
	\noindent\textbf{Visual Attributes Classification}: The first to investigate the power of visual attributes were Ferrari and Zisserman~\cite{ferrari2007learning}. They used low-level features and a probabilistic generative model to learn attributes of different types (\eg, appearance, shape, patterns) and segment them in an image. Kumar \etal~\cite{kumar} proposed an automatic method to perform face verification and image search. They first extracted and compared ``high-level'' visual features, or traits, of a face image that are insensitive to pose, illumination, expression, and other imaging conditions, and then trained classifiers for describable facial visual attributes (\eg, gender, race, and eyewear). A verification classifier on these outputs is finally trained to perform face verification. In the work of Scheirer \etal~\cite{scheirer2012multi}, raw attribute scores are calibrated to a multi-attribute space where each normalized value approximates the probability of that attribute appearing in the input image. This normalized multi-attribute space allows a uniform interpretation of the attributes to perform tasks such as face retrieval or attribute-based similarity search.
	Finally, attribute selection approaches have been introduced \cite{farhadi2009describing, wang2016category, zheng2016submodular} which select attributes based on specific criteria (\eg, entropy). Zheng \etal \cite{zheng2016submodular} formulated attribute selection as a submodular optimization problem \cite{krause2012submodular} and defined a novel submodular objective function. 
	
	Following the deep learning renaissance in 2012, several papers \cite{li2016human, rstarcnn, sarfraz2017deep, zhu2017learning, dong2017class} have addressed the visual attribute classification problem using ConvNets. Part-based methods decompose the image to parts and train separate networks which are then combined at a feature level before the classification step. They tend to perform well since they take advantage of spatial information (\eg, patches that correspond to the upper body can better predict the t-shirt color than others that correspond to other body parts). Zhang \etal~\cite{zhang2014panda} proposed an attribute classification method which combines part-based models in the form of poselets \cite{bourdev2011describing}, and deep learning by training pose-normalized ConvNets. Gkioxari \etal~\cite{gkioxari2015actions} proposed a deep version of poselets to detect human body parts which were then employed to perform action and attribute classification. Zhu \etal~\cite{zhu2015multi} introduced a method for pedestrian attribute classification. They proposed a ConvNet architecture comprising 15 separate subnetworks (\ie, one for each task) which are fed with images of different body parts to learn jointly the visual attributes. However, their method assumes that there is a pre-defined connection between parts and attributes and that all tasks depend on each other and thus, learning them jointly will be beneficial. Additionally, they trained the whole ConvNet end-to-end despite the fact that the size of the training dataset used was only 632 images. Based on our experiments, the only way to avoid heavy overfitting in datasets of that size is by employing a pre-trained network along with fine-tuning of some layers. Recycling pre-trained deep learning models with transfer learning (\ie, exploiting the discriminative power of a network trained for a specific task for a different problem or domain) is commonly used in the literature with great success \cite{sharif2014cnn, yosinski2014transferable}. Finally, visual attributes have been employed recently for re-identification \cite{su2018multi, su2017attributes}, pose estimation \cite{park2016attribute, sarafianos20163d}, 3D pose tracking \cite{livne2012human}, attribute mining and retrieval for clothing applications \cite{huang2015cross, song2016deep}, zero-shot visual object categorization and recognition \cite{lampert2014attribute} and image annotation and segmentation \cite{shi2016weakly}. 
	
	\noindent\textbf{Curriculum Learning}: Solving all tasks jointly is commonly employed in the literature \cite{ciregan2012multi, hand2016attributes, meyerson2017beyond} as it is fast, easy to scale, and achieves good generalization. For an overview of deep multi-task learning techniques, the interested reader is encouraged to refer to the work of Ruder \cite{ruder2017overview}. However, some tasks are easier than others and also not all tasks are equally related to each other \cite{pentina2015curriculum}. Curriculum Learning was initially proposed by Bengio \etal~\cite{bengio2009curriculum}. They argued that instead of employing samples at random it is better to present samples organized in a meaningful way so that less complex examples are presented first. Pentina \etal~\cite{pentina2015curriculum} introduced a curriculum learning-based approach to process multiple tasks in a sequence and developed a method to find the best order in which the tasks need to be learned. They proposed a data-dependent solution by introducing an upper-bound of the average expected error and employing an Adaptive SVM~\cite{yang2007cross, sarafianos2017adaptive}. Such a learning process has the advantage of exploiting prior knowledge to improve subsequent classification tasks but it cannot scale up to many tasks since each subsequent task has to be learned individually. Curriculum learning has also been employed with success on performing data regularization on models trained on corrupted labels \cite{jiang2017mentornet}, long short-term memory (LSTM) networks \cite{cirik2016visualizing, zaremba2014learning}, reinforcement learning \cite{tscl, florensa2017reverse}, robot learning policies \cite{murali2017cassl} as well as object detection\cite{li2017multiple}. In parallel with our work, Dong \etal~\cite{dong2016multi} also proposed a multi-task curriculum transfer technique to classify clothes based on their attributes. They approached the problem in a domain adaptation setup in which a classifier is first learned on easy clean samples (source domain) and then it is adapted to harder samples (cross-domain). However, the curriculum they utilize (which images correspond to the source domain and which to the cross-domain) is selected manually based on the dataset whereas in our proposed framework it is done automatically based on the label-cross correlation before training starts.
	
	\noindent\textbf{Transfer Learning}: Deep transfer learning techniques learn feature representations, which are transferable to other domains, by incorporating the adaptation to a new domain in the end-to-end learning process \cite{long2016deep, bengio2012deep}. The idea of distilling knowledge in neural networks was initially introduced by Hinton \etal~\cite{hinton2015distilling}. The authors proposed a method to distill the knowledge of a complex ensemble of models into a smaller model. The softmax output of the ensemble is divided by a temperature parameter and the smaller model learns directly from that ``softened'' output. Following that idea, Zhang \etal~\cite{zhang2016real} suggested a technique to perform action recognition in real-time. They transferred knowledge from the teacher (an optical flow ConvNet) to the student (a motion vector ConvNet) by backpropagating the teacher's loss in the students' network. Finally, Lopez-Paz \etal~\cite{LopSchBotVap16} introduced generalized distillation; a method that unifies the LUPI framework with the knowledge distillation paradigm.
	
	Finally, a very interesting prior work which focuses on the correlation of visual attributes is the method of Jayaraman \etal~\cite{jayaraman2014decorrelating}. Aiming to decorrelate attributes at learning time, the authors proposed a multi-task learning framework with the property of resisting the urge of sharing image features of correlated attributes. Their approach disambiguates attributes by isolating distinct low-level features for distinct properties (\eg, color for ``brown'', texture for ``furry''). They also leveraged side information for properties that are closely related and should share features (\eg, ``brown'' and ``red'' are likely to share the same features). While our work also leverages information from correlated attributes in a multi-task classification framework, it models co-occurrence between different clusters of visual attributes instead of trying to semantically decorrelate them. 
	
	\begin{figure*}[t] 
		\centering
		\includegraphics[width=0.96\textwidth]{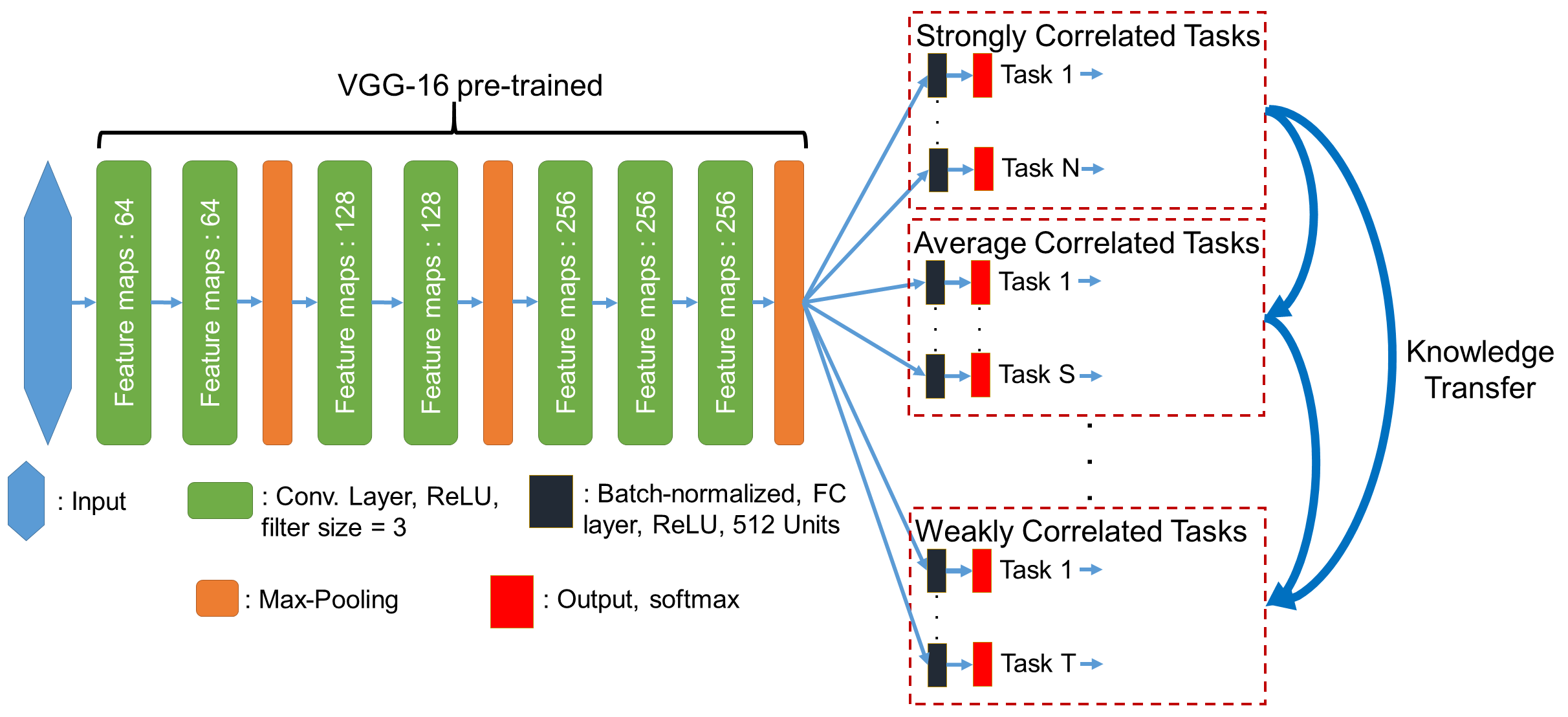}
		\caption{Architecture of the ConvNet used in our framework for several groups of tasks. The VGG-16 pre-trained part is kept frozen during training and only the weights of the last layers are learned. The different groups of tasks are learned sequentially using a curriculum learning paradigm. However, when the latter groups of tasks are trained, the tasks which have already been learned, contribute to the total cost function (Figure best viewed in color.)}
		\label{fig:Method}
	\end{figure*}
	
	\section{Methodology}\label{sec:method}
	In this section, we describe the proposed network architecture which given images of humans as an input, outputs visual attribute predictions. We then introduce our approach for splitting attributes into clusters. Finally, the proposed multi-task curriculum learning framework is introduced. 
	
	In our supervised learning paradigm, we are given tuples \((x_i,y_i)\) where \(x_i\) corresponds to images and \(y_i\) to the respective visual attribute labels. The total number of tasks will be denoted by \(T\), and thus the size of \(y_i\) for one image will be \(1\times T\). Finally, we will refer to the part of the network that solves the \(i^{th}\) group of tasks as \(C_i\). 	
	
	\subsection{Multi-label ConvNet architecture}
	To mitigate the lack of training data we employ the pre-trained VGG-16 \cite{simonyan2014very} network.  VGG-16, is the network from Simonyan and Zisserman which was one of the first methods to demonstrate that the depth of the network is a critical component for good performance. We selected VGG instead of a more modern network for the reason that it is a simple and homogeneous architecture, which despite its inefficiencies (\eg, large number of parameters), is sufficient for solving multiple binary image classification tasks.  VGG-16 is trained on ImageNet \cite{russakovsky2015imagenet}, the scale of which enables us to perform transfer learning between ImageNet and our tasks of interest. The architecture of the network we use is depicted in Figure~\ref{fig:Method}. We used the first seven convolutional layers of the VGG-16 network and dropped the rest of the convolutional and fully-connected layers. The reason behind this is that the representations learned in the last layers of the network are very task dependent \cite{yosinski2014transferable} and thus, not transferable. Following that, for every task we added a batch-normalized \cite{ioffe2015batch} fully-connected layer with 512 units and a ReLU activation function. We employed batch-normalization since it enabled higher learning rates, faster convergence, and reduced overfitting. Although shuffling and normalizing each batch has proven to reduce the need of dropout, we observed that adding a dropout layer \cite{srivastava2014dropout} was beneficial as it further reduced overfitting. The dropout probability was 75\% for datasets with less than 1,000 training samples and 50\% for the rest. For every task, an output layer is added with a softmax activation function using the categorical cross entropy.
	
	Furthermore, we observed that the random initialization of the parameters of the last two layers backpropagated large errors in the whole network even if we used different learning rates throughout our network. To address this behavior of the network, which is thoroughly discussed in the method of Sutskever \etal~\cite{sutskever2013importance}, we ``freeze'' the weights of the pre-trained part and train only the last two layers for each task in order to learn the layer weights and the parameters of the batch-normalization. 
	
	After we ensured that we can always overfit on the training set, which means that our network is deep enough and discriminative enough for the tasks of interest, our primary goal was to reduce overfitting. Towards this direction, we (i) selected 512 units for the fully connected layer to prevent the network from learning several weights; (ii) employed a small weight decay of \(10^{-4}\) for the layers that are trained; (iii) initialized the learning rate at \(10^{-3}\) and reduced it by a factor of 5 every 100 epochs and up to five times in total; and (iv) augmented the data by performing random scaling up to 150\% of the initial image followed by random crops, horizontal flips and adding noise by applying PCA to the RGB pixel values as proposed by Krizhevsky \etal~\cite{krizhevsky2012imagenet}. At test time, we averaged the predictions at three different scales (100\%, 125\%, and 150\%) of five fixed crops and their horizontal flips (30 in total) to obtain the predicted class label. This technique, which was also adopted in the ResNet method of He \etal \cite{resNets2016}, proved to be very effective as it reduced the variation on the predictions.
	
	\begin{figure*}[t] 
		\centering
		\begin{subfigure}[b]{0.52\textwidth}
			\includegraphics[width=\textwidth]{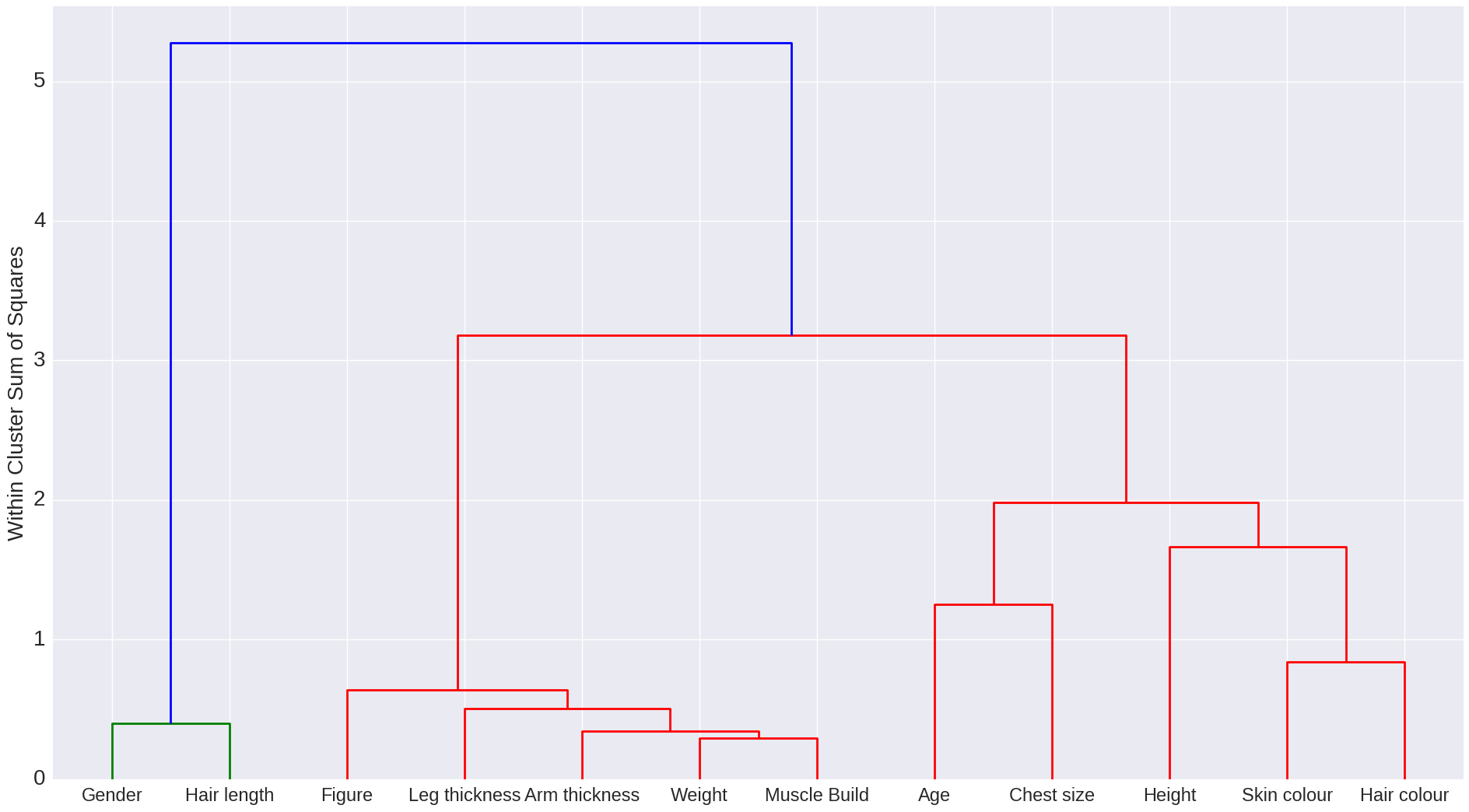}
		\end{subfigure}
		~ %add desired spacing between images, e. g. ~, \quad, \qquad, \hfill etc. 
		\begin{subfigure}[b]{0.46\textwidth}
			\includegraphics[width=\textwidth]{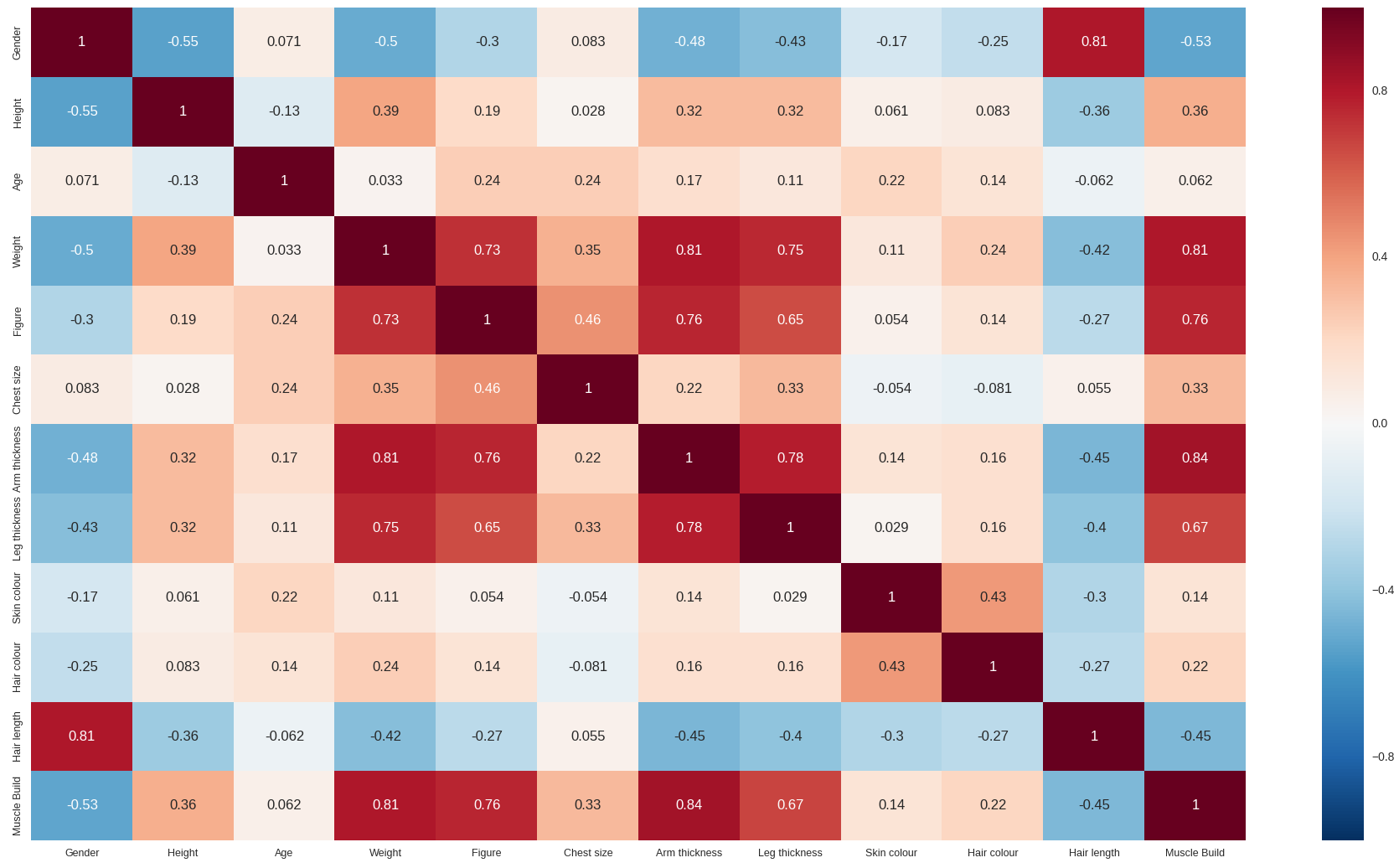}
		\end{subfigure}
		\caption{Dendrogram illustrating the arrangement of clusters (left) and the pairwise correlation matrix, which is fed to the clustering algorithm (right) of the visual attributes from the SoBiR dataset~\cite{martinho2016soft}. The sequence in which the clusters of attributes are learned, is obtained by computing the total dependency of each task with the rest within its cluster using Eq. (\ref{eu_eqnP}). The curriculum learning of the clusters of visual attributes is then performed in a descending order as described in Section~\ref{ssec:mtcur}. (Figure best viewed in color.)}
		\label{fig:hierchAndConfMat}
	\end{figure*}
	
	\subsection{Group Split with Hierarchical Clustering}\label{ssec:groupSplit}
	Finding the order in which tasks need to be learned so as to achieve the best performance is difficult and computationally expensive. Given some tasks \(t_i, i=1...T\) that need to be performed, we seek to find the best order in which the tasks should be performed so the average error of the tasks is minimized: 
	\begin{equation}\label{eq:1}
		\displaystyle \underset{\substack{S(t_i)}}{\text{minimize}} \; {\frac{1}{T}\sum\limits_{j=1}^{T} \mathcal{E}(\hat{y}_{t_j},y_{t_j}),}\\
	\end{equation}
	where \(S(t_i)\) is the function that finds the sequence of the tasks, \(\hat{y}_{t_j},y_{t_j}\) are the prediction and target vectors for the \(j^{th}\) task, and \(\mathcal{E}\) is the prediction error. 
	
	However, the fact that a task can be easily performed does not imply that it is positively correlated with another and that by transferring knowledge the performance of the latter will increase. Adjeroh \etal~\cite{D_2010_11241} studied the correlation between various anthropometric features and demonstrated that some correlation clusters can be derived in human metrology, whereby measurements in a cluster tend to be highly correlated with each other but not with the measurements in other clusters. 
	
	In this work, we seek to find: (i) which tasks (\ie, attributes) should be grouped together so as to be learned jointly, and (ii) which is the best sequence in which the groups of tasks should be learned. We use the training labels \(Y\) of size \(N\times M\) where \(N\) the number of samples, and \(M\) the number of attributes (\ie, ground truth labels) to compute the Pearson correlation coefficient matrix which is of size \(M\times M\). Each element in this matrix, represents to what extent these two attributes are correlated (\eg, the ``gender'' with the ``hair length'' will have a higher value compared to ``gender'' with ``age''). 
	
	We then employ the computed Pearson correlation coefficient matrix to perform hierarchical agglomerative clustering using the Ward variance minimization algorithm. Ward's method is biased towards generating clusters of the same size and analyzes all possible pairs of joined clusters, identifying which joint produces the smallest within cluster sum of squared (WCSS) errors. It is a variance-minimizing approach and which resembles the k-means algorithm but tackled with an agglomerative hierarchical approach. Assume that at an intermediate step, clusters \(s\) and \(t\) are to be merged to form cluster \(u=s\cup t\). Then, the new distance \(d(u,v)\) between cluster \(u\) and an already existing (but yet unused) cluster \(v\) is defined as: 
	\begin{equation} \label{eu_eqnWard}
		d(u,v) = \sqrt{\frac{|v| + |s|}{T}d(v,s)^2 + \frac{|v| + |t|}{T}d(v,t)^2 + \frac{|v|}{T}d(s,t)^2},
	\end{equation}
	where \(s\), \(t\) are the clusters which are joined into cluster \(u\), and \(T = |v| + |s| + |t|\). Ward~\cite{ward1963hierarchical}, points out that this procedure facilitates the identification of that union which has an objective function value ``equal or better than'' any of the \(n(n-1)/2\) possible unions. An illustrative hierarchical clustering of the visual attributes from the SoBiR dataset \cite{martinho2016soft} in the form of a dendrogram is depicted in Figure~\ref{fig:hierchAndConfMat}. We observe that the proposed method for task split yields clusters of visual attributes which cohere with our semantic understanding and intuition about which attributes might be related to each other (\eg, gender with hair length, weight with muscle build). In addition to the pairwise correlation matrix, which also provides an insight into the relation of attributes, the proposed approach exploits this correlation between the attributes during the learning process. 
	
	\begin{algorithm}[t]
		\SetKwInOut{Input}{Input}
		\SetKwInOut{Output}{Output}
		\Input{Training labels \(Y\), WCSS threshold \(\tau\)}
		\(P\leftarrow\) compute Pearson correlation coefficient matrix split based on labels \(Y\)\\
		\(G\leftarrow\) split into clusters using Eq. (\ref{eu_eqnWard}) along with \(P\), labels \(Y\), and \(\tau\)\\
		\For{group \(g_i\) in G}{
			\(S_i \leftarrow\) compute average of cross-correlation within \(g_i\) using Eq. (\ref{eu_eqnP}) \\    
		}
		\(S(g_i)\leftarrow\) compute learning sequence of clusters by sorting \(S_i\)'s in a descending order\\
		\Output{Learning sequence of clusters of visual attributes \(S(g_i)\)}
		\caption{Finding the learning sequence of attribute clusters}
		\label{alg2} 
	\end{algorithm}
	
	By splitting the attributes into clusters using a WCSS threshold \(\tau\) to cut the dendrogram horizontally, we have identified which tasks should be grouped together so as to be learned jointly. Following that, we now seek to obtain the sequence in which the clusters of visual attributes will be learned. To address this problem, we propose to find the total dependency \(p_{i,c}\) of task \(t_{i,c}\) with the rest within the cluster \(c\), by computing the respective Pearson correlation coefficients but this time only within the cluster as follows:  
	\begin{equation} \label{eu_eqnP}
		p_{i,c} =\sum_{j=1, j\neq i}^{T} \frac{cov(y_{t_{i,c}},y_{t_{j,c}})}{\sigma(y_{{t_{i,c}}})\sigma(y_{{t_{j,c}}})}, \; i=1,...,T
	\end{equation}
	where \(\sigma(y_{{t_{i,c}}})\) is the standard deviation of the labels \(y\) of the task \(t_{i,c}\). After we compute the total dependencies for all the clusters formed, we start the curriculum learning process in a descending order. 
	
	The process of computing the learning sequence of attribute clusters, which is described in detail in Algorithm.~\ref{alg2}, is performed once before the training starts. Since it only requires the training labels of the tasks to compute the cross-correlations and perform the clustering, it is not computationally intensive. Finally, note that the group split depends on the training set and it is possible that different train-test splits might yield different groups of tasks. 
	
	\subsection{Multi-Task Curriculum Learning}\label{ssec:mtcur}
	In the scenario we are investigating, we solve multiple binary unbalanced classification tasks simultaneously. Throughout the paper, we use the terms multi-label \cite{wang2016cnn} and multi-task interchangeably.  This is because we solve multiple classification tasks at the same time (multi-task), and at the same time for each given image we predict multiple binary labels which are not mutually exclusive (multi-label).
	
	\begin{algorithm}[t]
		\SetKwInOut{Input}{Input}
		\SetKwInOut{Output}{Output}
		\Input{Training set \(X\), training labels \(Y\), learning sequence of clusters \(S(g_i)\) from Algorithm~\ref{alg2}}
		\For{group \(g_i\) in \(S(g_i)\)}{
			Initialize \(C_i\) from rest of already trained groups of tasks (if any)\\
			\(C_i \leftarrow\) train model using \((X, Y_i)\) by minimizing the loss in Eq. (\ref{eu_eqn3}) \\    
		}    
		\Output{Parameters of network containing all groups of tasks}
		\caption{Multi-task curriculum learning training}
		\label{alg1} 
	\end{algorithm} 
	
	The proposed learning paradigm is described in Algorithm~\ref{alg1}. Similar to Zhu \etal~\cite{zhu2016multi}, we employ the categorical cross-entropy function between predictions and targets, which for a single attribute \(t\) is defined as follows: 
	\begin{equation} \label{eu_eqn1}
		L_t = -\frac{1}{N}\sum_{i=1}^{N}\sum_{j=1}^{M} \Bigg( \frac{\nicefrac{1}{M_j}}{\sum_{n=1}^{M}\nicefrac{1}{M_n}}\Bigg)\cdot \mathds{1}[y_{i}=j] \cdot \log(p_{i,j}),
	\end{equation}
	where \(\mathds{1}[y_{i}=j]\) is equal to one when the ground truth of sample \(i\) belongs to class \(j\), and zero otherwise, \(p_{i,j}\) is the respective prediction, which is the output of the softmax nonlinearity of sample \(i\) for class \(j\), and the term inside the parenthesis is a balancing parameter required due to imbalanced data. The total number of samples belonging to class \(j\) is denoted by \(M_j\), \(N\) is the number of samples and \(M\) the number of classes. The total loss over all attributes is defined as \(\sum_{t=1}^{T}\lambda_t \cdot L_t\), where \(\lambda_t\) is the contribution weight of each parameter. For simplicity, it is set to \(\lambda_t = \nicefrac{1}{T}\). By setting \(\lambda_t\) in this way, there is an underlying assumption that all tasks contribute equally to the multi-task classification problem. To overcome this limitation, a fully-connected layer with \(T\) units could be added with an identity activation function after each separate loss \(L_t\) is computed. In that way, the respective weight for each attribute in the total loss function could be learned. However, we observed that for groups of tasks that consist of a few attributes there was no difference in the performance, and thus we did not investigate this any further. 
	
	Once the classification of the visual-attribute tasks that demonstrated the strongest intra correlation is performed, we use the learned parameters (\ie, weights, biases, and batch normalization parameters) to initialize the network for the less diverse groups of attributes. The architecture of the network remains the same, with the parameters of VGG-16 being kept ``frozen''. The weights of the tasks of previous groups of clusters continue to be learned with a very small learning rate of \(10^{-6}\)). Furthermore, by adopting the ``supervision transfer'' technique of Zhang \etal~\cite{zhang2016real} we leverage the knowledge learned by backpropagating the following loss: 
	\begin{equation} \label{eu_eqn3}
		L_j = \lambda\cdot L_t + (1-\lambda)\cdot L_j^f,
	\end{equation}
	where \(L_j^f\) is the total loss computed during the forward pass using Eq. (\ref{eu_eqn1}) over only the current group of correlated tasks and \(\lambda\) is a parameter that controls the amount of knowledge transferred. Since the parameters of the network that correspond to already trained groups of tasks keep being updated, the loss \(L_t\) changes during training of the tasks of interest each time. This enables us to transfer the knowledge from groups of tasks with stronger intra cross-correlation to groups which demonstrated less intra cross-correlation. This technique proved to be very effective, as it enhanced the performance of the parts of the network which are responsible for the prediction of less correlated groups of tasks, and contributed to faster convergence during training.  
	
	%		\begin{figure}[t] 
	%			\centering
	%			\includegraphics[width=0.45\textwidth]{figures/multiClass.png}
	%			\caption{Throughout the paper, we use the terms multi-label and multi-task interchangeably. CILICIA is a multi-task classification approach since it learns multiple tasks in a joint manner. However, since it solves multiple binary classification tasks at once it is also a multi-label classification paradigm.}
	%			\label{fig:multiClass}
	%		\end{figure}		
	
	\section{Experiments}\label{sec:exp}    
	\subsection{Datasets}
\begin{figure}[t] 
		\centering
		\includegraphics[width=0.46\textwidth]{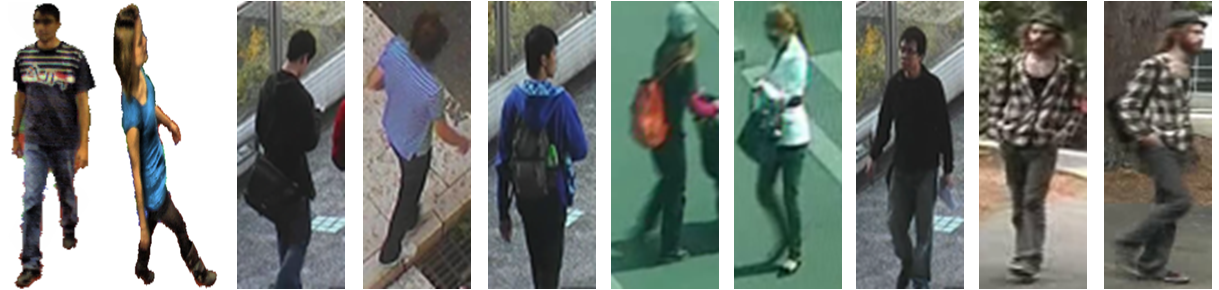}
		\caption{Example images of pedestrians captured from different viewpoints. Attributes comprise backpacks, hats, clothing color or patterns and soft-biometrics such as gender or age.}
		\label{fig:datasetExamples}
	\end{figure}

	To verify the effectiveness of the proposed method, we conducted evaluations in three challenging datasets containing standing humans, and thus tested our method in almost all the possible variations that can be found in the datasets used in the literature. We used the SoBiR \cite{martinho2016soft}, VIPeR \cite{gray2007evaluating} and PETA \cite{deng2014pedestrian} datasets. The selected datasets are of varying difficulty and contain different visual attributes and training set sizes. In each dataset we follow the same evaluation protocol with the rest of the literature. Some representative images are depicted in Figure~\ref{fig:datasetExamples}. 
	
	\noindent\textbf{SoBiR dataset}: The recently introduced SoBiR dataset \cite{martinho2016soft} contains 800 images of 100 people. The dimensions of each image are \(256\times256\). The SoBiR dataset comprises 12 soft biometric labels (\eg, gender, weight, age, height) and four forms of comprehensive human annotation (absolute versus relative and categorical versus binary). In our experimental investigation, we used the comparative binary ground-truth annotations (\eg, taller/shorter instead of tall/short) instead of absolute binary. The main reasons for this choice are: (i) relative binary annotations have been shown to outperform categorical annotations \cite{martinho2016soft, parikh2011relative}; and (ii) class labels were balanced for all soft biometrics. A 80/10/10 train/validation/test split based on human IDs is performed (so that only new subjects appear at testing) and average classification results are reported over five random splits.
	
	\noindent\textbf{VIPeR dataset}: The VIPeR dataset \cite{gray2007evaluating} contains 1,264 low-resolution (\(128 \times 48\)) pedestrian images. Each individual is captured in just a pair of images from a different camera, under different viewpoint, pose and lighting conditions. Layne \etal~\cite{layne2014attributes} provided 21 visual-attribute annotations which are used in our evaluation. We randomly split VIPeR into non-overlapping training and testing sets of equal sizes based on the human IDs. Following the literature, we repeated this process six times (one split for parameter tuning and the rest for evaluation of our method) and average classification results are reported.
	
	\noindent\textbf{PETA dataset}: The PETA dataset \cite{deng2014pedestrian} consists of 19,000 images gathered from 10 different smaller datasets. Parameters such as the camera angle, viewpoint, illumination, and resolution are highly variant, which makes it a valuable dataset for visual-attribute classification evaluation. It is divided in 9,500, 1,900, and 7,600 images for training, validation, and testing, respectively. Similar to \cite{zhu2016multi}, highly imbalanced attributes are discarded and the remaining 45 binary visual attributes are employed.
	
	\subsection{Results on SoBiR}
	\noindent\textbf{Implementation details}: For the SoBiR dataset, the batch size was set to 160. We split it into four clusters containing 2, 5, 2, and 3 attributes by thresholding at within cluster sum of squares \(\tau = 1.9\) (Figure~\ref{fig:hierchAndConfMat}), trained our models for 5,000 epochs, and set \(\lambda=0.25\). 
	
\begin{figure}[t] 
		\centering
		\includegraphics[width=0.48\textwidth]{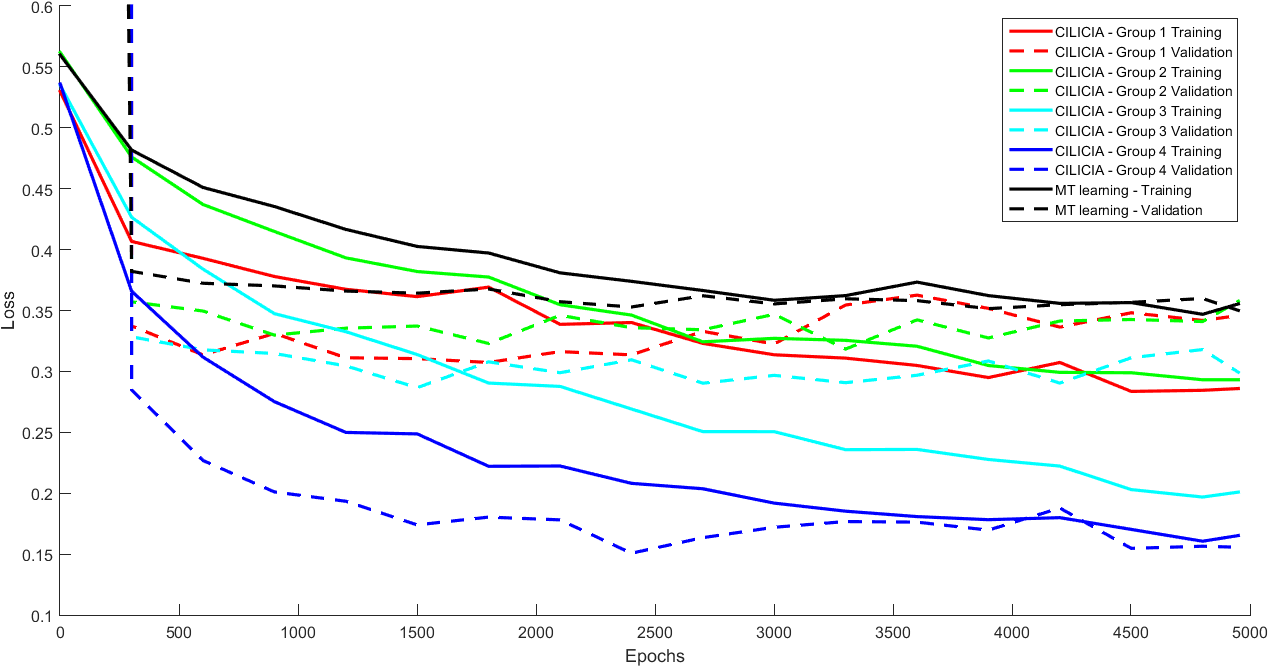}
		\caption{Convergence plot for all the groups of CILICIA as well as Multi-Task learning on the SoBiR dataset. Note that the numbering starts from the most correlated to the least correlated cluster. (Figure best viewed in color.)}
		\label{fig:converg}
	\end{figure}

	\noindent\textbf{Evaluation results}: Since the SoBiR dataset does not have a baseline on attribute classification we reported results using handcrafted features and an SVM classifier as well as three different end-to-end learning frameworks using our ConvNet architecture. In all cases, images were resized to \(128\times 128\). The features used for training the SVMs consisted of: (i) edge-based features, (ii) local binary patterns (LBPs), (iii) color histograms, and (iv) histograms of oriented gradients (HOGs). To preserve local information, we computed the aforementioned features in four blocks for every image resulting in 540 features in total. In addition, we performed SVM with features extracted from the last fully-connected layer of the pre-trained VGG-16 network and the obtained results are provided in the third column of Table~\ref{tab:SoBiR}. Feature vectors \(4,096 \times 1\) were extracted for each image, and an SVM was trained using the optimal parameters obtained from the validation set. Furthermore, we investigated the classification performance when tasks are learned individually (\ie, by backpropagating only their own loss in the network), jointly in a typical multi-task classification setup (\ie, by backpropagating the average of the total loss in the network), and using the proposed approach.  We report the classification accuracy (\%) for all 12 soft biometrics in Table~\ref{tab:SoBiR}. CILICIA is superior in both groups of tasks to the rest of the learning frameworks. Despite the small size of the dataset, ConvNet-based methods perform better in all tasks compared to an SVM with handcrafted features. Multi-task learning methods (\ie, multi-task and CILICIA) outperform the learning frameworks when tasks are learned independently since they leverage information from other attributes. By taking advantage of the correlation between attributes, CILICIA demonstrated higher classification performance than a typical multi-task learning scenario. However, estimating the ``age'' proved to be the most challenging task in all cases as its classification accuracy ranges from 58.5\% to 64.5\% when it is learned individually using our ConvNet architecture. This poor performance can be attributed to the fact that age estimation from images without facial traits is a largely unsolved problem \cite{xing2017diagnosing, li2017d2c}. In Figure~\ref{fig:converg}, the convergence plots for all four CILICIA groups are depicted and the following observations are made: (i) the first group (comprises only two attributes) after epoch 3,000, demonstrates strong overfitting which proved to be inevitable even when we experimented with smaller learning rates; (ii) Multi-Task learning demonstrated the highest loss compared to the groups of the proposed method; and (iii) as we move from the groups of attributes that are strongly correlated to the rest by transferring knowledge each time, the training loss becomes smaller and there is less overfitting (if any). Note that the depicted losses for the corresponding groups are averaged over the tasks that belong to the cluster and thus, they can be compared although the number of tasks in each group is not the same. 

	\begin{table*}[t]
		\centering
		\caption{Classification accuracy of different learning paradigms on the SoBiR dataset. In individual learning, each attribute is learned separately. In multi-task learning, the average loss of all attributes is backpropagated in the network. In CILICIA, four clusters were formed and attributes are in descending order based on their intra cross-correlation. Results highlighted with \colorbox{blue!15}{light purple} indicate statistically significant improvement using the paired-sample t-test.} 
		\small    
		\begin{tabular}{l P{2.8cm} P{2cm} P{1.2cm} P{1.5cm} P{1.4cm}}
			\toprule
			Soft Label  &  SVM with Handcrafted Features &  SVM with Deep Features & Individual Learning & Multi-Task Learning & CILICIA\\
			\midrule
			Gender & 72.1 & 74.5 & 80.4 & 79.6 & \cellcolor{blue!15}\textbf{85.2} \\
			Height &  64.7 & 61.8 & 73.9 & 72.0 & \textbf{77.0} \\
			Age &  58.5 & 55.3 & 62.6 & 61.9 & \textbf{64.5} \\
			Weight & 57.7 & 65.3 & 67.7 & 71.0 & \cellcolor{blue!15}\textbf{74.1} \\
			Figure &  57.8 & 64.3 & \textbf{68.7} & 67.1 & 67.3 \\
			Chest size &  58.7 & 54.5 & 64.9 & \textbf{68.9} & 67.5 \\
			Arm thickness & 60.1 & 70.5 & 72.0 & 73.1 & \cellcolor{blue!15}\textbf{73.7} \\
			Leg thickness & 56.7 & 65.5 & 68.9 & 71.0 & \cellcolor{blue!15}\textbf{72.6} \\
			Skin color & 59.2 & 54.3 & 66.8 & 67.6 & \textbf{68.7} \\
			Hair color &  67.5 & 59.5 & 74.2 & 76.1 & \textbf{77.9} \\
			Hair length & 71.8 & 72.5 & 78.9 & 79.2 & \cellcolor{blue!15}\textbf{85.9} \\
			Muscle build & 58.5 & 66.3 & 73.3 & 74.5 & \cellcolor{blue!15}\textbf{75.8} \\
			\midrule
			Average &  61.9 & 63.6 & 71.0 & 71.3 & \textbf{74.2}\\
			\bottomrule
		\end{tabular}%
		\label{tab:SoBiR}%
	\end{table*}%
	
\begin{figure*}[t] 
		\centering
		\begin{subfigure}[b]{0.48\textwidth}
			\includegraphics[width=\textwidth]{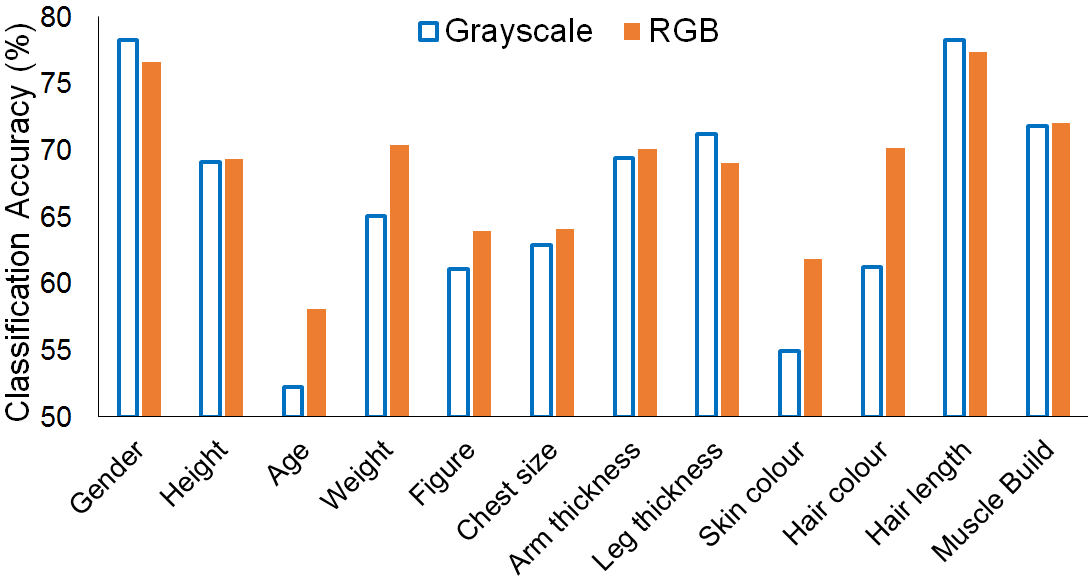}
		\end{subfigure}
		~ %add desired spacing between images, e. g. ~, \quad, \qquad, \hfill etc. 
		\begin{subfigure}[b]{0.49\textwidth}
			\includegraphics[width=\textwidth]{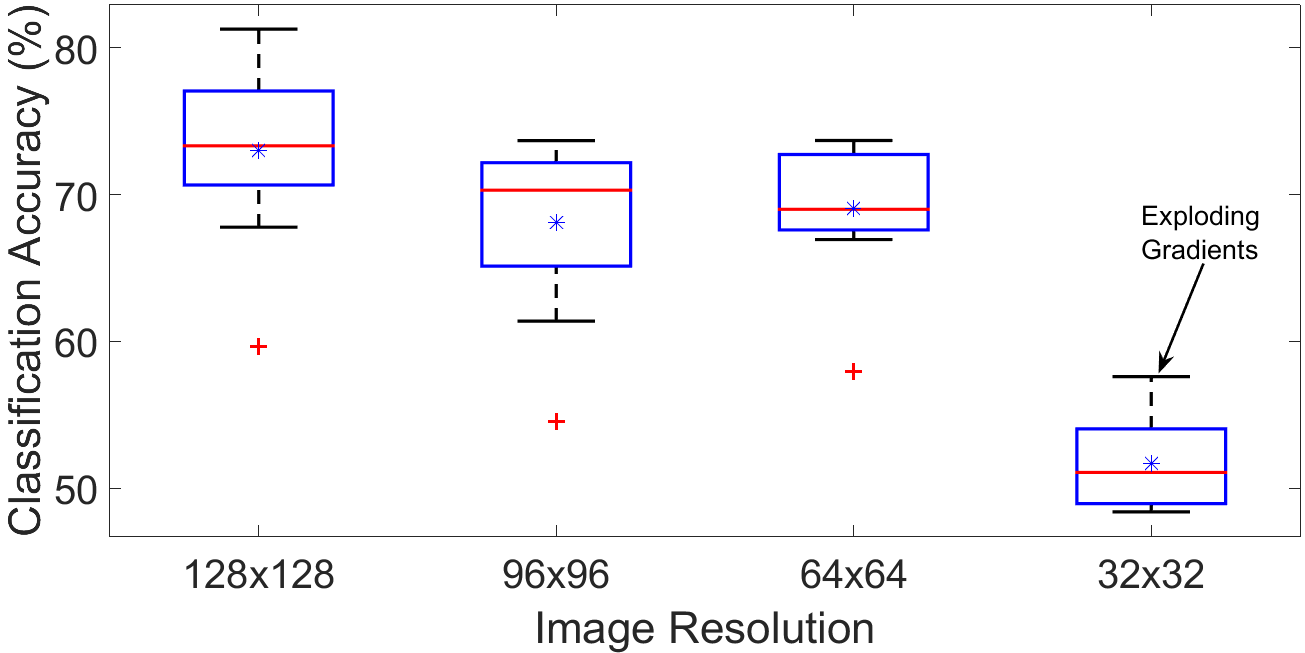}
		\end{subfigure}
		\caption{Impact of image color information in the classification accuracy of the attributes of the SoBiR dataset (left) and classification accuracy for different image resolutions (right). The standard deviation is high because we averaged over all attributes of the SoBiR dataset. Outliers depicted with a red cross correspond to ``age''.}
		\label{fig:grayrgb}
	\end{figure*}

	\noindent\textbf{Impact of color information}: In this experiment, we investigated to what extent the color information affects the classification accuracy of the visual attributes. Since we employed a network pre-trained on RGB images, we fed each of the three color channels with the same grayscale images. Figure~\ref{fig:grayrgb} (left) summarizes the obtained results. Visual attributes such as hair and skin color were expected to have a drop in performance when color information was absent. However, we observed that weight and age were also affected as their accuracy dropped by 8\% and 11\% respectively. 
	
	%	\begin{figure}[t] 
	%		\centering
	%		\includegraphics[width=0.55\textwidth]{figures/grayrgb2.png}
	%		\caption{Impact of image color information in the classification accuracy of the attributes of the SoBiR dataset.}
	%		\label{fig:grayrgb}
	%	\end{figure}
	%	
	%	\begin{figure}[t] 
	%		\centering
	%		\includegraphics[width=0.55\textwidth]{figures/boxplot.png}
	%		\caption{Classification accuracy for different image resolutions. The standard deviation is high because we averaged over all attributes of the SoBiR dataset. Outliers depicted with a red cross correspond to ``age''.}
	%		\label{fig:boxplot}
	%	\end{figure}
	
	\noindent\textbf{Impact of image resolution}: The objective of this experiment is to assess the impact of image resolution on the classification performance. One of the major advantages of ConvNets (due to parameter sharing) is that they do not need to have a fixed-size input and pre-trained networks can be utilized with images of different spatial size. This still holds for fully-connected layers since they are equivalent to convolution layers with \(1\times 1\) kernel. Our ConvNet architecture was fed with images of varying resolutions starting from \(32\times 32\) up to \(128\times 128\) and average classification accuracies over all attributes are reported. From the obtained results in Figure~\ref{fig:grayrgb} (right), we draw the following observations. First, images of higher resolution tend to perform better than lower since they provide more spatial space for the convolutional operations. Second, when we experimented with images of size \(32\times 32\), we observed that the norm of the gradient started taking very high values which is a common phenomenon during training ConvNets and is referred to as the exploding gradients problem \cite{pascanu2013difficulty}. In our case, the reasons for the explosion of gradients were the small image size in a network pre-trained on images of almost 10-times larger and the existence of dropout with a high probability before the output layer.
	\setlength{\tabcolsep}{2.5pt}
	\setlength{\textfloatsep}{0.8cm}
	\begin{table}[t]
		\centering
		\caption{Performance comparison on the VIPeR dataset. Five clusters were formed and attributes are in descending order based on their intra cross-correlation. Results highlighted with \colorbox{blue!15}{light purple} indicate statistically significant improvement using the z-test.}
		\footnotesize    
		\begin{tabular}{p{2.1cm} P{1.5cm} P{2.3cm} c}
			\toprule
			Visual Attribute  & Multi-Task Learning & Zhu \etal~\cite{zhu2015multi} & CILICIA\\
			\midrule
			barelegs & 79.6 $\pm$ 0.8 & 84.1 $\pm$ 1.1 & \cellcolor{blue!15}\textbf{88.6} $\pm$ 0.4\\
			shorts& 76.8 $\pm$ 1.1 & 81.7 $\pm$ 1.3 & \cellcolor{blue!15}\textbf{89.6} $\pm$ 0.6\\
			lightshirt& 79.5 $\pm$ 0.9 & 83.0 $\pm$ 1.2 & \cellcolor{blue!15}\textbf{84.6} $\pm$ 0.6\\
			nocoats& 74.3 $\pm$ 1.3 & 71.3 $\pm$ 0.8 & \cellcolor{blue!15}\textbf{75.0} $\pm$ 0.3 \\
			blueshirt& 69.9 $\pm$ 1.7 & 69.1 $\pm$ 3.3 & \cellcolor{blue!15}\textbf{90.2} $\pm$ 1.2 \\
			midhair& 74.3 $\pm$ 1.3 & 76.1 $\pm$ 1.8 & \textbf{77.7} $\pm$ 0.6\\
			lightbottoms& \textbf{79.0} $\pm$ 1.0 & 76.4 $\pm$ 1.2 & 75.1 $\pm$ 1.2 \\
			redshirt& 79.2 $\pm$ 1.9 & 91.9 $\pm$ 1.0 & \cellcolor{blue!15}\textbf{93.6} $\pm$ 0.4 \\
			nolightdarkjeans& 87.1 $\pm$ 1.6 & 90.7 $\pm$ 2.0 & \cellcolor{blue!15}\textbf{96.0} $\pm$ 0.5\\
			greenshirt& 70.3 $\pm$ 2.4 & 75.9 $\pm$ 5.9 & \cellcolor{blue!15}\textbf{95.1} $\pm$  0.4\\
			hashandbag& 66.9 $\pm$ 3.1 & 42.0 $\pm$ 6.5 & \cellcolor{blue!15}\textbf{90.8} $\pm$ 0.6\\
			hassatchel& 72.5 $\pm$ 0.8 & 57.8 $\pm$ 2.7 & \cellcolor{blue!15}\textbf{72.8} $\pm$ 0.4 \\
			skirt& 67.2 $\pm$ 3.7 & 78.1 $\pm$ 3.5 & \cellcolor{blue!15}\textbf{94.8} $\pm$ 0.6 \\    
			darkbottoms& 68.1 $\pm$ 0.9 & \textbf{78.4} $\pm$ 0.7 & 76.3 $\pm$ 0.9\\    
			male& 71.5 $\pm$ 1.9 & 69.6 $\pm$ 2.6  & \cellcolor{blue!15}\textbf{81.0} $\pm$ 1.9\\
			patterned& 67.4 $\pm$ 3.5 & 57.9 $\pm$ 9.2 & \cellcolor{blue!15}\textbf{92.2} $\pm$ 0.7 \\
			darkshirt& 71.0 $\pm$ 1.4 & 82.3 $\pm$ 1.4 & \cellcolor{blue!15}\textbf{84.3} $\pm$ 0.5\\
			jeans& 74.9 $\pm$ 0.7 & \textbf{77.5} $\pm$  0.6 & 76.2 $\pm$ 0.6\\        
			darkhair& 70.1 $\pm$ 2.0 & \textbf{73.1} $\pm$ 2.1 & 71.8 $\pm$ 1.3\\            
			hasbackpack& 68.4 $\pm$ 1.4 & 64.9 $\pm$ 1.2 & \cellcolor{blue!15}\textbf{76.1} $\pm$ 1.6\\
			\midrule
			Total Av. & 73.4 $\pm$ 1.2 & 74.1 $\pm$ 1.0 & \textbf{84.0} $\pm$ 0.8 \\
			\bottomrule
		\end{tabular}%
		\label{tab:VIPeR}%
	\end{table}%
	
	\subsection{Results on VIPeR}
	\noindent\textbf{Implementation details}: For the VIPeR dataset, the batch size was set to 158. We split it into five clusters containing 2, 4, 7, 3, and 4 attributes by thresholding at within cluster sum of squares \(\tau = 1.8\), trained our models for 5,000 epochs, and set \(\lambda=0.25\).         
	
	\begin{figure}[t] 
		\centering
		\includegraphics[width=0.48\textwidth]{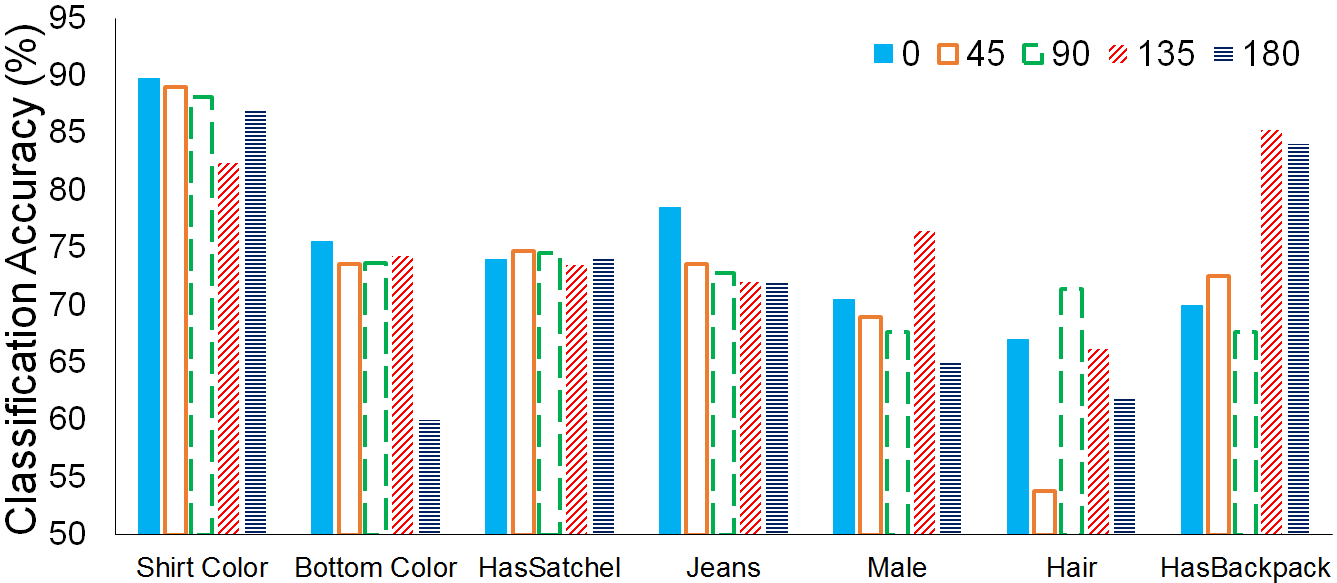}
		\caption{Classification performance for the low-correlated attributes under different viewpoints. For visualization purposes, we group attributes (\eg, ``darkhair'' and ``midhair'' correspond to ``hair'').}
		\label{fig:boxplot1}
	\end{figure}
	
	\noindent\textbf{Evaluation results}: To demonstrate the superiority of the proposed approach over normal multi-task learning approaches, we evaluate in Table~\ref{tab:VIPeR} its performance in comparison with the method of Zhu \etal~\cite{zhu2015multi} and a typical multi-task learning framework. To test for statistical significance between CILICIA and the method of Zhu \etal~\cite{zhu2015multi} we employed the z-test (\(p < 0.05\)) since the mean and the standard deviation results were available from their technique. Employing the proposed multi-task curriculum learning approach is beneficial for the classification of visual attributes, as it outperformed the previous state-of-the-art by improving the total results by \(9.9\%\). CILICIA achieved significantly better results in most of the tasks, which demonstrates the efficacy of our method over traditional multi-task learning approaches. The reason for this is that when some tasks are completely unrelated then multi-task learning has a negative effect as it forces the network to learn representations that explain everything, which is not possible. Additionally, we observed that color attributes tend to achieve higher performance compared to other attributes. The reason for this is that such attributes are highly imbalanced (sometimes more than one to nine) due to the way annotation is provided (\eg, is the human wearing a red t-shirt or not). Note that, since in both the SoBiR and the VIPeR datasets, the training set is not fixed and train-test splits are performed each time, the sequence in which the groups of tasks are learned is not necessarily the same each time. The reason for this is that the hierarchical agglomerative clustering depends on the pairwise correlation matrix between the labels of the training set which vary between different splits. 
	
	\noindent\textbf{Impact of facial information}: To investigate the impact of facial information in the classification, we trained the proposed architecture with images of humans after removing the upper part of the image containing the face (top 20\% of the image). From the results in Table~\ref{tab:noFace}, we observed that the performance of the visual attributes related to clothes or objects of the upper body was not affected by the absence of facial information. On the contrary, the impact on the hair-related attributes was significant since their performance dropped by 6\%. Finally, facial information plays a vital role in recognizing the gender of humans even in low-resolution images as a 5.3\% performance drop was observed in the absence of the face. 
	
	\begin{table}[b]
		\centering
		\caption{Performance comparison of the weakly correlated visual attributes when the face exists and it is absent in the input image. Attributes are grouped based on where they correspond to the human body.}
		\small    
		\begin{tabular}{l c c c c}
			\toprule
			& Upper Body & Lower Body & Hair & Gender \\
			\midrule
			Entire Image & \textbf{85.3} & 84.8 & \textbf{74.7} & \textbf{81.0} \\
			No Face &  83.2 & \textbf{85.4} & 68.7 & 75.7  \\
			\bottomrule
		\end{tabular}%
		\label{tab:noFace}%
	\end{table}% 
	
	\noindent\textbf{Impact of viewpoint}: Aiming to investigate to what extent the camera viewpoint affects the classification performance, we report classification results over the weakly-correlated attributes depending on the camera angle. VIPeR images are captured from \(0^o, 45^o, 90^o, 135^o, \) and \(180^o\) degrees and the obtained results are depicted in Figure~\ref{fig:boxplot1}. It can be observed that visual attributes related to shirt color are viewpoint-invariant, whereas others such as ``jeans'' perform better from a frontal angle. Two interesting observations arise from the hair and backpack attributes. First, classifying attributes pertaining to hair (length and color) can be done with higher accuracy when the viewpoint is at \(90^o\). Second, finding whether the human has a backpack or not becomes an easier task for camera angles of \(135^o\) and \(180^o\) which is compatible with the way humans would perform on this task.         
	
	\begin{figure*}[t] 
		\centering
		\includegraphics[width=0.94\textwidth]{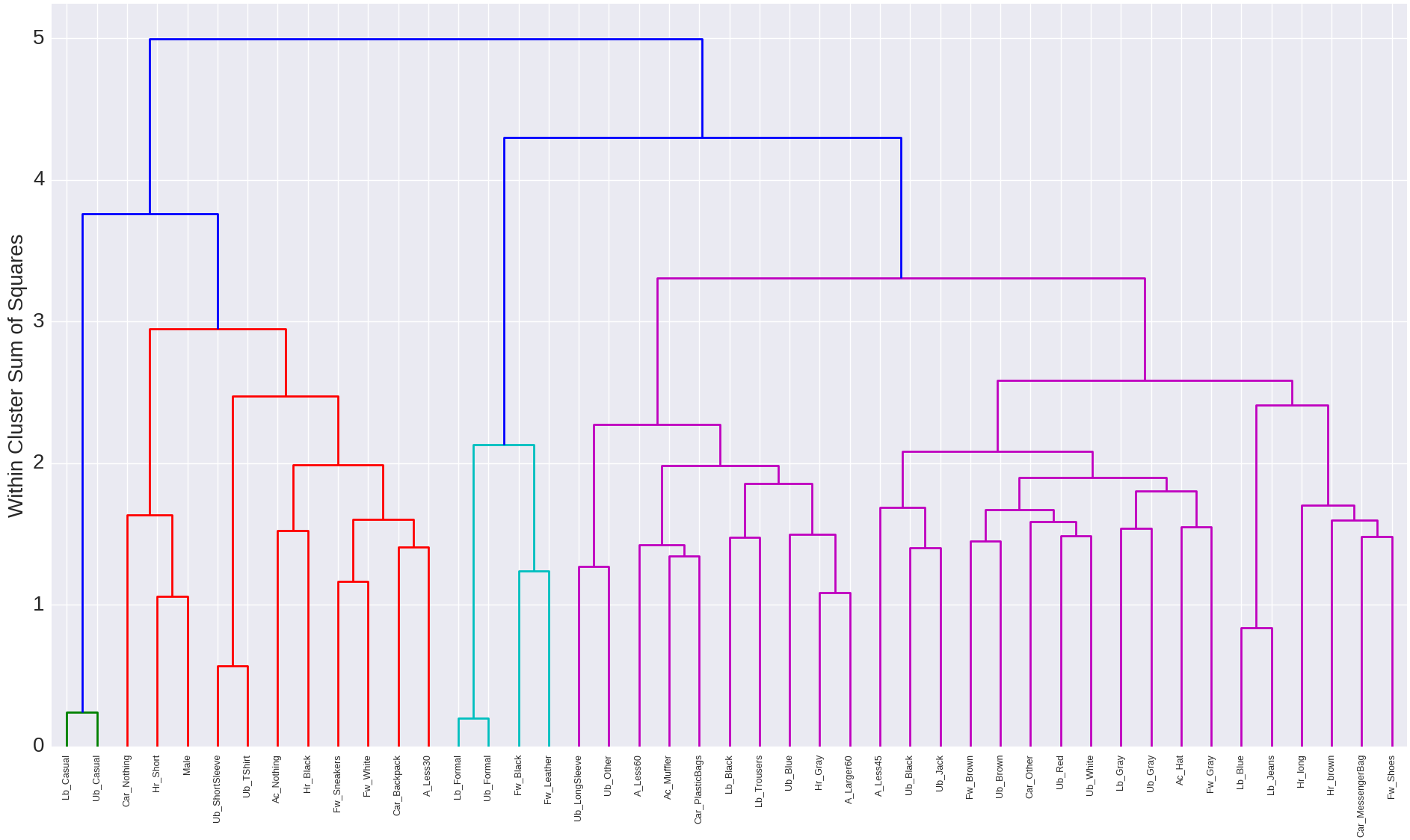}
		\caption{Dendrogram of the visual attributes of the PETA training set. For WCSS equal to 3, five clusters are formed. The learning sequence of the clusters is green, left purple, red, right purple and turquoise. 
			(Lb. - Lower Body, Ub. - Upper Body, Car. - Carrying, Ac. - Accessory, Fw. Footwear, Hr. - Hair, and A. - Age ).}
		\label{fig:PetaTree}
	\end{figure*}

	\subsection{Results on PETA}
	\noindent\textbf{Implementation details}: For the PETA dataset, the batch size was set to 190. We split it into five clusters containing 2, 11, 4, 10, and 18 attributes by thresholding at within cluster sum of squares \(\tau = 3\), trained our models for 5,000 epochs, and set \(\lambda=0.2\). 
	
	\begin{figure}[t] 
		\centering
		\includegraphics[width=0.48\textwidth]{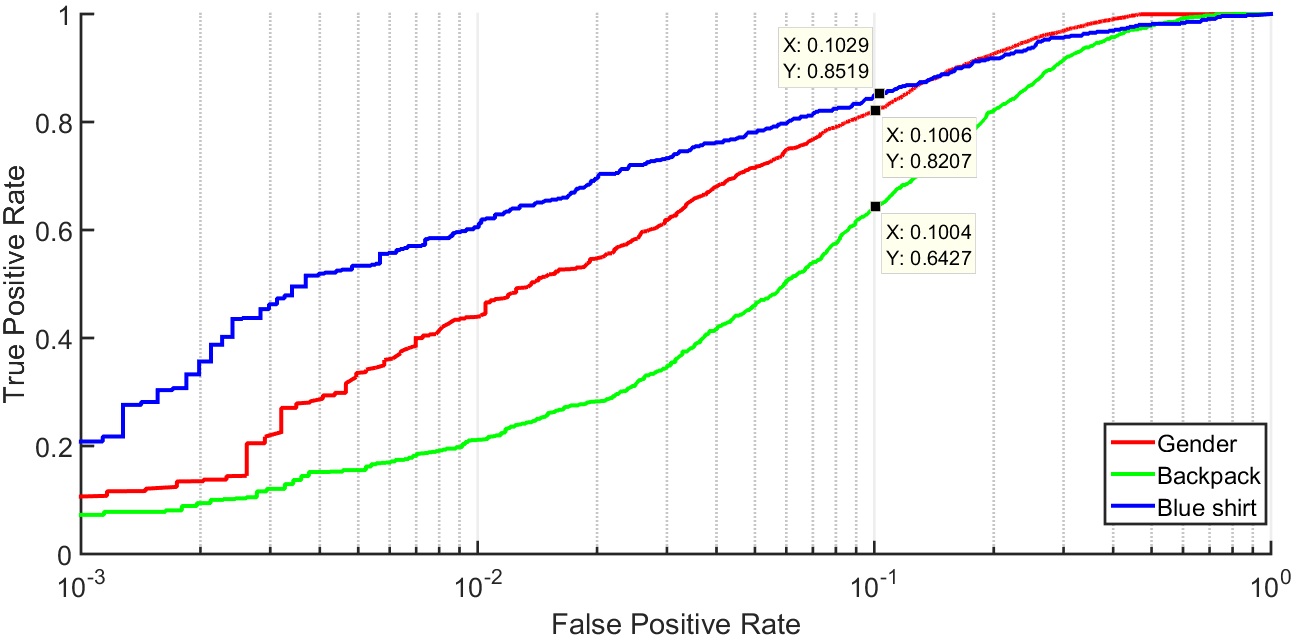}
		\caption{ROC curves for the visual attributes of ``gender'', ``blue shirt'', and ``has backpack''. The x-axis is in semi-logarithmic scale and the depicted values correspond to the recall rate (\%) when the false positive rate is 10\%.}
		\label{fig:Roc}
	\end{figure}
	
	\noindent\textbf{Evaluation results}: Since the training size of the PETA dataset is significantly higher than the rest (almost 10,000) and the annotations provided are 45 instead of 20, some very interesting observations can be made from the clusters of visual attributes depicted in Figure~\ref{fig:PetaTree}. The turquoise cluster comprises attributes related to upper and lower body formal clothes along with black and leather footwear, and thus it is beneficial if we learn these attributes at the same time. Other examples that follow to our intuition and semantic understanding are the fact that being male is very strongly connected with having short hair and not carrying any type of bag, or that carrying a backpack is linked with being less than 30 years old. The proposed learning approach employs this information from attributes strongly connected on the PETA dataset and outperformed the recent method of Zhu \etal~\cite{zhu2016multi}. 
	
	Since many attributes are highly imbalanced and the classification accuracy as an evaluation metric is not sufficient by itself they also reported recall rate results when the false positive rate is equal to 10\% as well as the area under the ROC curve (AUC). Following the same evaluation protocol, we tested the proposed multi-task curriculum learning method on the PETA dataset and report our results in comparison with those of Zhu \etal~\cite{zhu2016multi} after grouping the attributes in Table~\ref{tab:petaResults1}. Although our method is not part-based, as it does not split the human image into parts which are then learned individually, it outperforms the part-based method of Zhu \etal~\cite{zhu2016multi} in all types of visual attributes under all evaluation metrics. Due to highly imbalanced data (the imbalance ratio in most of the categories is relatively high), the improvement in the classification accuracy is minor. However, for the rest of the evaluation metrics, our method improved the average recall rate by 3.93\% and the AUC by 1.94\%. In Figure~\ref{fig:Roc} the ROC curves of some tasks in which our method performed really well (\eg, ``blue shirt''), reasonably well (\eg,``gender''), and adequately (\eg,``has backpack'') are depicted. The complete results on the PETA dataset are provided in Table~\ref{tab:petaResults}.
	
	\begin{table*}[t]
		\centering
		\caption{Performance comparison on the PETA dataset for different types of attributes. The imbalance ratio is defined as the ratio of the number of instances in the majority class to the number of examples in the minority class in the training set.}
		\small    
		\begin{tabular}{l c c c c c c c}
			\toprule
			\multirow{2}{*}{Visual Attribute} & \multirow{2}{*}{Imbalance Ratio} & \multicolumn{2}{c}{Accuracy (\%)} & \multicolumn{2}{c}{Recall rate (\%) @FPR = 10\%} & \multicolumn{2}{c}{AUC (\%)} \\ 
			\cmidrule(r){3-4} \cmidrule(r){5-6} \cmidrule(r){7-8}
			
			& & Zhu \etal~\cite{zhu2016multi} & CILICIA & Zhu \etal~\cite{zhu2016multi} & CILICIA & Zhu \etal~\cite{zhu2016multi} & CILICIA\\
			\midrule
			Accessories & 7.63 & 93.11 & \textbf{93.48} & 75.68 & \textbf{76.66} & 91.06 & \textbf{92.13}\\ 
			Carrying Bags & 5.01 & 83.68 & \textbf{84.78} & 57.21 & \textbf{62.73} & 82.79 & \textbf{85.63} \\ 
			Footwear & 4.69 & 83.41 & \textbf{83.74} & 59.09 & \textbf{60.88} & 84.44 & \textbf{85.18} \\ 
			Hair & 4.57 & 89.54 & \textbf{89.96} & 75.89 & \textbf{80.43} & 90.95 & \textbf{93.18} \\ 
			Lower Body & 3.54 &  85.05 & \textbf{85.66} & 64.92 & \textbf{66.95} & 87.37 & \textbf{88.26}\\ 
			Upper Body & 8.06 & 89.60 & \textbf{90.48} & 69.88 & \textbf{76.12} & 88.66 & \textbf{91.68} \\ 
			Age & 7.05 & 87.84 & \textbf{87.90} & 71.03 & \textbf{72.49} & 88.93 & \textbf{90.24}\\ 
			Gender & 1.22 & 84.34 & \textbf{87.59} & 74.80 & \textbf{82.04} & 91.74 & \textbf{93.84} \\ 
			\midrule
			Total Av.  & 6.07 & 87.23 & \textbf{87.91} & 67.29 & \textbf{71.22} & 87.66 & \textbf{89.60} \\  
			\bottomrule
		\end{tabular}%
		\label{tab:petaResults1}%
	\end{table*}%            

	\section{Ablation Studies and Performance Analysis}\label{sec:ablations}
	\subsection{Is Hierarchical Clustering Beneficial?}
\begin{figure}[t] 
		\centering
		\includegraphics[width=0.48\textwidth]{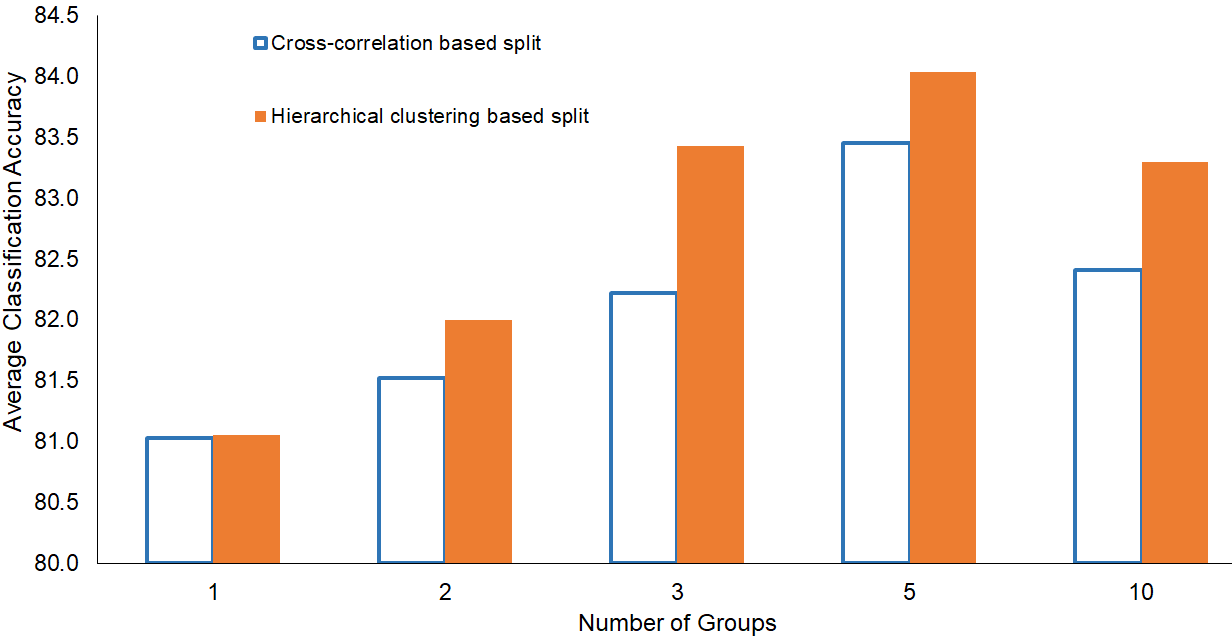}
		\caption{Average classification accuracy over all visual attributes of the VIPeR dataset, for different number of group splits. The cross-correlation based split refers to grouping the tasks based on their total cross-correlation with the rest after arranging them in a descending order. The hierarchical clustering based split corresponds to the proposed approach described in Algorithm \ref{alg2}.}
		\label{fig:hierSplit}
	\end{figure}

	An important question that arises while analyzing the performance of CILICIA on attribute classification is, what is the impact of the group split using hierarchical clustering along with the proposed learning paradigm which guides the training process? To what extent can the obtained results be accustomed to using a deep-learning multi-task classification scheme? To answer such questions, we conducted a detailed experimental evaluation on the VIPeR dataset \cite{gray2007evaluating} for a different number of groups that were split with two different methods. The first group split is performed after sorting the total cross-correlations for each attribute using Eq. (\ref{eu_eqnP}), in descending order, and splitting the attributes to a fixed number of groups. For example, for the VIPeR dataset, we compute for each attribute the sum of cross-correlations with the rest, which results in a \(20\times 1\) vector. After sorting this vector, we split it into \(N\) number of groups depending on the number of groups that we are interested in investigating. The second way to split the visual attributes into groups is by employing the proposed approach (described in Section~\ref{ssec:groupSplit}). In Figure~\ref{fig:hierSplit}, we demonstrate the average classification accuracy over all visual attributes of the VIPeR dataset \cite{gray2007evaluating} for a different number of groups, using the two methods described. We observe that splitting into groups using a hierarchical bottom-up method yields better results in all groups over the cross-correlation bases split and that for the VIPeR dataset, five clusters are the optimal number of groups. The reason it works better than simply grouping the attributes based on the cross-correlation, is because attributes are assigned to clusters by identifying which joint produces the smallest WCSS of errors.
	
	\subsection{Why is Knowledge Transfer important?}\label{ssec:whykt}
	To assess the impact of transferring knowledge from groups of tasks which have already converged to ones that have not been learned yet we conducted an ablation experiment. We selected the four most correlated and the four least correlated attributes of the PETA dataset so as to form the two groups of strongly and weakly correlated attributes. We compare the classification accuracy of the selected tasks with and without knowledge transfer. When no knowledge is transferred to the latter group, we are simply training two multi-task classification frameworks. We report the obtained results in the last two columns of Table~\ref{tab:Abl1}. Transferring knowledge from a strongly correlated group of tasks to the weakly improves the performance of the latter by 1.89\% compared to a typical multi-task classification learning framework. 
	
	\subsection{Why use Correlation as a Criterion for Group Split?}
	To demonstrate the effectiveness of clustering attributes into groups based on their cross correlation we conducted an ablation study using the same eight attributes (as in Section~\ref{ssec:whykt}) from the PETA dataset. However, in this experiment, instead of grouping them based on their cross-correlation, we randomly assign them to two groups. We follow exactly the same two-stage process (\ie, learning one group first and transferring knowledge to the second which is learned right after) and report the obtained results in the first column of Table~\ref{tab:Abl1}. We observe that learning in correlation-based groups of tasks is beneficial as CILICIA with and without knowledge transfer performs better than learning at random. Additionally, transferring knowledge between attributes that do not co-occur (or they are semantically completely different) has an adverse effect on the performance. The obtained results are in line with previous methods that can be found in the literature \cite{huang2012multi, hariharan2010large} that have exploited label correlations to improve multi-task learning.
	
	\subsection{Why is the Proposed Curriculum the Right One?}
	
	We argue that task similarity and thus the curriculum is not binary, but resides on a spectrum. In the same way that humans learn with different curricula depending on the task, the process of finding a curriculum that is beneficial for all tasks cannot have an optimal single solution. Learning in correlation-split groups showed promising results (Tables~\ref{tab:SoBiR} and \ref{tab:Abl1}) which led us to start considering how can we improve the performance. Transferring knowledge between related tasks is not beneficial as during the joint multi-task learning training the parameter sharing plays that role. Transferring knowledge from randomly-split groups also proved to be ineffective (Table~\ref{tab:Abl1}). We then investigated whether the work of Bengio \etal~\cite{bengio2009curriculum}, which proposed a curriculum based on what is easier to learn first, would add value. We believe that the knowledge transfer from the strongly to the weakly correlated group of tasks is a reasonable easy-to-hard curriculum which resembles to the definition of Bengio \etal~\cite{bengio2009curriculum}. In addition, note that when Bengio \etal~\cite{bengio2009curriculum} introduced curriculum learning after they defined an entropy-based curriculum they demonstrated that introducing gradually more difficult examples speeds-up online training. In our paper, this can be observed in the convergence plot (Figure~\ref{fig:converg}) in which the subgroups converge faster and with a smaller loss (average among tasks).

	\begin{table}[t]
		\centering
		\caption{Ablation experiments to assess the effectiveness of knowledge transfer and correlation-based split. In the random split column, the strongly and weakly groups refer only to the learning sequence as the split is not based on the correlation. CILICIA (w/o kt) corresponds to learning in correlation-split groups but without knowledge transfer.}
		\small    
		\begin{tabular}{l c c c}
			\toprule
			Group  & Random Split & CILICIA (w/o kt) & CILICIA\\
			\midrule
			Strongly & 65.36 & 76.01 & 76.01 \\
			Weakly & 63.08 & 69.91 & \textbf{71.80}\\
			Total & 64.22 & 72.95 & \textbf{73.91}\\
			\bottomrule
		\end{tabular}%
		\label{tab:Abl1}%
	\end{table}% 
	
	\subsection{Performance Analysis and Limitations}
	The proposed approach outperformed the state-of-the-art in all datasets and demonstrated better results over the rest of the learning paradigms on the SoBiR dataset. The main reasons for this are: (i) we exploited the correlation between different attributes and split them into clusters using hierarchical clustering; (ii) we proposed a learning paradigm to learn the group of attributes in a curriculum learning framework and classify them in a multi-task classification setup; and (iii) we leveraged the already learned clusters of visual attributes which had converged to transfer knowledge to groups that were about to be learned to improve the performance and enhance the stability of our method. 
	
	Despite its success and good performance, the proposed approach has a few limitations and inefficiencies. First, the existence of a fully-connected layer after the last convolutional layer increases significantly the number of parameters that need to be learned for each task. We partially addressed this by freezing most of the network and employed a small number of units in the fully-connected layer. This inefficiency is known for the VGG network and was addressed by more recent networks that such as the GoogLeNet \cite{szegedy2015going}, the ResNet \cite{resNets2016} or the Highway Networks \cite{srivastava2015highway}. To investigate to what extent this inefficiency affects the overall performance we evaluated the performance of the ResNet-18 network against VGG-16 by fine-tuning both networks end-to-end on the VIPeR dataset. Following a multi-task classification setup using all \(20\) attributes we observed that using the ResNet-18 results in \(77.93\%\) classification accuracy which is 4\% higher than the corresponding accuracy of VGG. This result demonstrates the importance of residual connections that prevent gradients from vanishing while training deep neural networks. Second, the proposed approach contains two additional parameters that need to be cross-validated thoroughly. The first parameter is \(\lambda\), which controls the contribution of the already learned groups of clusters and is found in several methods that perform transfer learning or knowledge distillation \cite{LopSchBotVap16, zhang2016real}. For this parameter, we experimented on the validation set with different parameters (namely 0.25, 0.5, 0.75 and 1) and observed that a 25\% contribution of the already learned clusters of visual attributes was the most effective. The second parameter is the within-cluster sum of squares threshold which controls the number of clusters formed. Finally, the goal of the proposed approach was to classify the visual attributes of humans, the full body of whom was always fully-visible. Thus, it was tested in re-identification datasets, which contain pairs of images of humans standing or walking, and outperformed the state-of-the-art without even following a part-based approach. For datasets such as the Berkeley Attributes of People dataset \cite{bourdev2011describing}, which comprises humans of varying poses with parts of their body either not visible or occluded, part-based (or poselet-based) approaches \cite{gkioxari2015actions, rstarcnn, li2016human} have proven to be very effective recently. To address these challenges, CILICIA would have to be adapted to work with poselets or body-parts (\ie, in which order the body parts need to be learned so as to transfer information between groups of tasks) which was outside the scope of this paper.
	
	%        \begin{figure}[t] 
	%            \centering
	%            \includegraphics[width=0.45\textwidth]{figures/datasetExamples1.png}
	%            \caption{Examples of correct and incorrect attribute predictions.}
	%            \label{fig:datasetExamples1}
	%        \end{figure}

	\section{Conclusion}\label{sec:conc}
	Given a set of tasks that need to be learned we sought to find an answer to how we can learn them effectively and what would be the optimal way in terms of performance, speed, and simplicity. Learning each task separately, although very simple, lacks in terms of performance since it does not exploit the information from other tasks.  Learning all tasks at the same time in a multi-task classification scenario is relatively fast, easy to implement, and employs knowledge from other tasks to boost the classification performance. Curriculum learning is a learning scheme in which samples or tasks are not treated as equally easy or hard, but are instead presented to the model in a meaningful way so as to increase generalization and performance. Since learning a large number of tasks one at a time is computationally expensive, we opted for learning clusters of tasks in a curriculum. In each cluster of visual attributes, we proposed to learn the corresponding tasks in a multi-task classification setup. 
	
	Our proposed method, CILICIA, finds the sequence in which clusters of visual attributes are learned very efficiently, and classifies them with high performance. Given images of standing humans as an input, we performed end-to-end learning by solving multiple binary classification problems simultaneously. Tasks were grouped into clusters by employing hierarchical agglomerative clustering based on their correlation. The sequence (\ie, curriculum) in which clusters were learned was found by computing the average cross-correlation within each cluster and sorting the obtained values in a descending order. During training of weakly correlated clusters of tasks, we leveraged the knowledge already learned from clusters which demonstrated stronger correlation. By these means, we combined the advantages of both multi-task and curriculum learning paradigms; since our method converges fast, it is effective and employs prior knowledge. We evaluated our method in three datasets and outperformed the state-of-the-art by 9.9\% on the VIPeR dataset and by a recall rate of almost 4\% (when the false positive rate is fixed and equal to 10\%) on the PETA dataset despite the fact that no body part-specific information was employed. The obtained results demonstrate the effectiveness and, at the same time, the great potential of multi-task curriculum learning. 
	
	\section*{Acknowledgments}
	This work has been funded in part by the UH Hugh Roy and Lillie Cranz Cullen Endowment Fund. The work of C. Nikou is supported by the European Commission (H2020-MSCA-IF-2014), under grant agreement No 656094. All statements of fact, opinion or conclusions contained herein are those of the authors and should not be construed as representing the official views or policies of the sponsors.
	
	\section*{References}
	\bibliographystyle{elsarticle-num} 
	\bibliography{Refs}
	\clearpage
	\begin{table*}[t!]
		\centering
		\caption{Performance comparison on the PETA dataset for different types of attributes. The imbalance ratio is defined as the ratio of the number of instances in the majority class to the number of examples in the minority class in the training set. An asterisk next to an attribute denotes that it belongs to the strongly correlated group of tasks.}
		\resizebox{\textwidth}{!}{
			\begin{tabular}{l c c c c c c c}
				\toprule
				\multirow{2}{*}{Visual Attribute} & \multirow{2}{*}{Imbalance Ratio} & \multicolumn{2}{c}{Accuracy (\%)} & \multicolumn{2}{c}{Recall rate (\%) @FPR = 10\%} & \multicolumn{2}{c}{AUC (\%)} \\ 
				\cmidrule(r){3-4} \cmidrule(r){5-6} \cmidrule(r){7-8}
				
				& & Zhu \etal~\cite{zhu2016multi} & CILICIA & Zhu \etal~\cite{zhu2016multi} & CILICIA & Zhu \etal~\cite{zhu2016multi} & CILICIA\\
				\midrule
				accessoryHat*& \(\,\,\, 8.78\) & 96.05 & \textbf{96.56} & \textbf{86.06} & 85.23 & \textbf{92.62} & 92.28\\ 
				accessoryMuffler*& \(11.06\) & \textbf{97.17} & 97.04 & 88.42 & \textbf{91.23} & 94.47& \textbf{95.62}\\ 
				accessoryNothing& \(\,\,\, 3.03\) & 86.11& \textbf{86.45} & \textbf{52.57} & 51.93 & 86.09& \textbf{86.75} \\ 
				\midrule
				carryingBackpack& \(\,\,\, 4.01\) & 84.30 & \textbf{84.82} & 58.40 & \textbf{64.20} & 85.19& \textbf{89.62}\\ 
				carryingMessengerBag& \(\,\,\, 2.37\) & 79.58& \textbf{80.59} & 58.30 &\textbf{60.13} & 82.01&\textbf{83.35} \\ 
				carryingNothing& \(\,\,\, 2.68\) & 80.14& \textbf{81.12}& 55.15 & \textbf{61.11}& 83.08&\textbf{85.30} \\ 
				carryingOther& \(\,\,\, 4.11\) &80.91&\textbf{81.69} & 46.90 &\textbf{50.68} & 77.68&\textbf{78.16} \\ 
				carryingPlasticBags*& \(11.87\) & 93.45 & \textbf{94.72} & 67.30 & \textbf{76.11} & 86.01& \textbf{90.77}\\ 
				\midrule
				footwearBlack& \(\,\,\, 1.26\) & \textbf{75.97} & 75.33 & \textbf{57.24}& 55.59 &\textbf{84.07}& 83.42\\  
				footwearBrown*& \(13.64\) &92.14 & \textbf{92.65} & \textbf{65.77} &62.73  & 85.26&\textbf{85.28} \\  
				footwearGrey& \(\,\,\, 5.40\) &87.07& \textbf{88.07} & 50.80 & \textbf{60.54} & 80.92& \textbf{84.54}\\ 
				footwearLeatherShoes& \(\,\,\, 2.39\) &85.26 & \textbf{86.16} & 72.28 & \textbf{76.79} & 89.84 & \textbf{90.96}\\ 
				footwearShoes& \(\,\,\, 1.75\)& 75.78 &\textbf{76.95} & 52.80 &\textbf{55.52}& 81.63& \textbf{81.73}\\  
				footwearSneakers*& \(\,\,\, 3.67\) &\textbf{81.78} & 81.08& \textbf{52.04} & 50.63& 83.19&\textbf{83.43} \\  
				footwearWhite*& \(\,\,\, 4.73\) &\textbf{85.89}& 85.73& \textbf{62.72} &60.44 & \textbf{86.16}& 85.26\\ 
				\midrule 
				hairBlack& \(\,\,\, 1.55\)&87.83& \textbf{87.97} & 81.03 & \textbf{82.46} & 93.61&\textbf{94.13} \\  
				hairBrown& \(\,\,\, 3.86\) & \textbf{89.58} &89.07 & 77.36 & \textbf{81.22}& 91.33& \textbf{92.48}\\  
				hairGrey*& \(11.36\) &\textbf{95.25} & 95.09 & 74.91 & \textbf{80.65} & 89.42 & \textbf{91.65} \\  
				hairLong& \(\,\,\, 3.23\) & 88.12 & \textbf{88.21} & 76.49 & \textbf{78.93} & 90.55& \textbf{92.15}\\     
				hairShort*& \(\,\,\, 2.84\) & 86.93 & \textbf{87.03} & \textbf{69.68} & 69.28 & 89.84 & \textbf{90.38} \\ 	 
				\midrule
				lowerBodyBlack& \(\,\,\, 1.08\) & 83.86 & \textbf{85.41} & 71.21 & \textbf{79.42} & 90.84&\textbf{92.83} \\ 
				lowerBodyBlue*& \(\,\,\, 4.51\) & 88.64 & \textbf{89.53} & 77.26 & \textbf{80.16} & 90.81& \textbf{92.06} \\  
				lowerBodyCasual& \(\,\,\, 6.33\) & \textbf{90.54} & 89.58 & 56.23 & \textbf{58.76} & 87.49 & \textbf{88.19}\\
				lowerBodyFormal& \(\,\,\, 6.45\) & 90.86 & \textbf{90.87} & 72.52  & \textbf{72.94} & 87.79& \textbf{88.94}\\ 
				lowerBodyGrey*& \(\,\,\, 3.10\) & \textbf{82.07} & 81.85 & \textbf{53.43} & 51.73 & 82.77&\textbf{83.33} \\ 
				lowerBodyJeans*& \(\,\,\, 2.28\) & \textbf{83.13} & 82.57 & \textbf{67.59} & 64.85& \textbf{87.71}&86.46 \\  
				lowerBodyTrousers* & \(\,\,\, 1.06\) & \textbf{76.26}& 75.90 & \textbf{56.19} & 54.77 & \textbf{84.16} & 83.34\\  
				\midrule
				personalLarger60* & \(16.21\) & \textbf{97.58} & 97.24 & \textbf{90.71} & 90.33 & 94.94 & \textbf{95.96}\\  
				personalLess30 & \(\,\,\, 1.00\) & 81.05 & \textbf{81.84} & 63.75 & \textbf{67.34} & 88.50& \textbf{89.42}\\  
				personalLess45 & \(\,\,\, 2.03\)  & \textbf{79.87}  & 79.39 & \textbf{59.42} & 58.64  & 84.62& \textbf{85.25}\\  
				personalLess60* & \(\,\,\, 8.96\) & 92.84 & \textbf{93.21} & 70.22 & \textbf{71.94} & 87.66& \textbf{88.03}\\ 
				\midrule
				personalMale* & \(\,\,\, 1.22\) & 84.34 & \textbf{86.03} & 74.80 & \textbf{77.87} & 91.74 & \textbf{92.85}\\ 
				\midrule
				upperBodyBlack & \(\,\,\, 1.25\) & 86.21 & \textbf{86.86} & 80.11 & \textbf{82.07} & 93.06&\textbf{94.07} \\ 
				upperBodyBlue* & \(12.05\) & 94.53 & \textbf{95.74} & 76.19 & \textbf{84.64} & 90.92 & \textbf{94.51}\\  
				upperBodyBrown* & \(12.59\) & 93.25& \textbf{94.20} & 68.60 & \textbf{79.80} & 87.58& \textbf{92.54}\\ 
				upperBodyCasual* & \(\,\,\, 5.98\) & \textbf{89.25} & 80.00 & \textbf{62.14} & 61.45 & \textbf{87.17}& 85.08\\ 
				upperBodyFormal & \(\,\,\, 6.66\) & \textbf{91.12} & 90.33 & 70.48 & \textbf{74.47} & 87.57& \textbf{89.60}\\ 
				upperBodyGrey & \(\,\,\, 4.56\) & 84.39 & \textbf{84.46} & 55.33 & \textbf{63.38} & 82.99& \textbf{87.44}\\ 
				upperBodyJacket* & \(13.20\) & 92.34 & \textbf{93.04} & 53.37 & \textbf{63.10} & 80.98 & \textbf{85.27} \\  
				upperBodyLongSleeve & \(\,\,\, 4.98\) & 87.88 & \textbf{87.96} & 74.29 & \textbf{80.69 }& 89.97& \textbf{92.72} \\  
				upperBodyOther* & \(\,\,\, 1.22\) & \textbf{81.97} & 81.58& \textbf{73.19} & 71.83 &\textbf{88.50}& 88.14\\  
				upperBodyRed* & \(17.66\) & 96.33 & \textbf{96.64} & 86.77 & \textbf{90.82}& 94.69& \textbf{96.16}\\  
				upperBodyShortSleeve & \(\,\,\, 5.98\) & 88.09  &\textbf{88.79} & 69.22 & \textbf{75.88}& 89.21& \textbf{92.10}\\  
				upperBodyTshirt* & \(10.53\) & 90.59 & \textbf{90.80} & 63.51 & \textbf{68.65} & 88.73 & \textbf{90.55}\\ 
				upperBodyWhite & \(18.87\) & 88.84 & \textbf{90.37} & 75.25 & \textbf{84.18} & 91.24& \textbf{94.65}\\   
				\midrule
				Strongly Cor. Av. & \(\,\,\, 8.12\) & 89.63 & \textbf{89.87} & 70.04 & \textbf{72.19} & 88.42 & \textbf{89.50} \\  
				Weakly Cor. Av. & \(\,\,\, 4.12\) & 84.93 & \textbf{85.32} & 64.66 & \textbf{68.56} & 86.93 & \textbf{88.60} \\  
				Total Av.  & \(\,\,\, 6.07\) & 87.23 & \textbf{87.54} & 67.29 & \textbf{70.34} & 87.66 & \textbf{89.04} \\  
				\bottomrule
			\end{tabular}%
		}
		\label{tab:petaResults}%
	\end{table*}%	
	
\end{document}